%% file: neurips_2025.tex
\documentclass{article}

\usepackage[preprint]{neurips_2025}

\usepackage[utf8]{inputenc} % allow utf-8 input
\usepackage[T1]{fontenc}    % use 8-bit T1 fonts
\usepackage{hyperref}       % hyperlinks
\usepackage{url}            % simple URL typesetting
\usepackage{booktabs}       % professional-quality tables
\usepackage{amsfonts}       % blackboard math symbols
\usepackage{nicefrac}       % compact symbols for 1/2, etc.
\usepackage{microtype}      % microtypography

\usepackage{amsthm}
\usepackage{multirow}
\usepackage{array}
\usepackage{placeins}
\usepackage{adjustbox}
\usepackage{algorithmic}
\usepackage[table]{xcolor}
\definecolor{commentgray}{rgb}{0.5,0.5,0.5}
\definecolor{lightgray}{gray}{0.8}
\usepackage{amsmath,amssymb}
\usepackage[ruled,vlined]{algorithm2e}
\usepackage[skins]{tcolorbox}
\usepackage{titletoc}
\usepackage{empheq}
\usepackage{xspace}
\usepackage{float}
\usepackage{minitoc}
\usepackage{enumitem}
\usepackage{caption}
\usepackage{wrapfig}
\usepackage{tabularx}

\setlist[itemize]{topsep=-1pt, itemsep=1pt, parsep=0pt, partopsep=0pt,left=1pt}

\definecolor{rowgray}{gray}{0.96}

\SetKwSwitch{Switch}{Case}{Other}{switch}{do}{other}
\SetKwProg{Switch}{Switch}{}{}
\SetKwBlock{Initialize}{Initialize}{}

\SetKwFunction{Sample}{Sample}
\SetKwFunction{Grid}{Grid}
\SetKwFunction{Detect}{Detect}
\SetKwFunction{Verify}{Verify}
\SetKwFunction{Spline}{Spline}
\SetKwFunction{Gaussian}{Gaussian}
\SetKwFunction{TopK}{TopK}
\SetKwFunction{ComputeRelation}{ComputeRelation}
\SetKwFunction{Detected}{Detected}
\SetKwFunction{Normalize}{Normalize}

\newcommand{\fancyname}{\texttt{Visual Semantic-Logical Search}\xspace}
\newcommand{\fancynameshort}{\texttt{VSLS}\xspace}
\newcommand{\tstar}{\texttt{T\textasteriskcentered}\xspace}
\newcommand{\lvb}{\textsc{LongVideoBench}\xspace}
\newcommand{\videomme}{\textsc{Video-MME}\xspace}
\newcommand{\hlvb}{\textsc{Haystack-LVBench}\xspace}
\newcommand{\lvh}{\textsc{LV-Haystack}\xspace}
\title{Logic-in-Frames: Dynamic Keyframe Search \\ via Visual Semantic-Logical Verification \\ for Long Video Understanding}

\author{
Weiyu Guo \quad Ziyang Chen \quad Shaoguang Wang \quad Jianxiang He \quad Yijie Xu\\
AI Thrust, HKUST(GZ)\\
{\tt\small \{\href{mailto:wguo395@hkust-gz.edu.cn}{wguo395}, 
\href{mailto:zchen483@connect.hkust-gz.edu.cn}{zchen483}, 
\href{mailto:swang440@connect.hkust-gz.edu.cn}{swang440}, 
\href{mailto:jhe307@connect.hkust-gz.edu.cn}{jhe307}, 
\href{mailto:yxu409@connect.hkust-gz.edu.cn}{yxu409}\}@connect.hkust-gz.edu.cn}
\and
\textbf{Jinhui Ye}\\
Shanghai AI Laboratory\\
{\tt\small \href{mailto:jinhuiyes@gmail.com}{jinhuiyes@gmail.com}}
\and
\textbf{Ying Sun}\textsuperscript{\footnotemark[1] \ } \quad \textbf{Hui Xiong}\textsuperscript{\footnotemark[1] \ }\\
AI Thrust, HKUST(GZ)\\
{\tt\small \{\href{mailto:yings@ust.hk}{yings}, 
\href{mailto:xionghui@ust.hk}{xionghui}\}@ust.hk}
}

\begin{document}

\doparttoc
\faketableofcontents
\maketitle
\footnotetext[1]{Corresponding authors.}
\input{sec/abstract}
\input{sec/intro}

\input{sec/Method}
\input{sec/Experiment}
\input{sec/Analysis}

\input{sec/relatedwork}
\input{sec/conclusion}

\begingroup
\small
\renewcommand{\refname}{References}

\bibliographystyle{ieeenat_fullname}
\bibliography{neurips_2025}
\endgroup

%%%%%%%%%%%%%%%%%%%%%%%%%%%%%%%%%%%%%%%%%%%%%%%%%%%%%%%%%%%%
\newpage
\appendix
\addcontentsline{toc}{section}{Appendix}
\part{Appendix}
\parttoc
\newpage

\input{sec/appendix.tex}

%%%%%%%%%%%%%%%%%%%%%%%%%%%%%%%%%%%%%%%%%%%%%%%%%%%%%%%%%%%%

\newpage
\section*{NeurIPS Paper Checklist}

\begin{enumerate}

\item {\bf Claims}
    \item[] Question: Do the main claims made in the abstract and introduction accurately reflect the paper's contributions and scope?
    \item[] Answer: \answerYes{}
    \item[] Justification: The abstract and introduction clearly state the main contributions of our work, including (1) the proposal of a semantics-driven keyframe search framework using four logical relations, (2) performance gains on multiple long video QA benchmarks, (3) efficient frame sampling (1.4\%) with state-of-the-art results, and (4) plug-and-play compatibility with VLM/LLM pipelines. These claims are supported by both the method and experimental sections (see Sections “Introduction”, “Method”, and “Experiment”), and limitations are discussed in the main paper and Appendix~\ref{app:limitations}. The claims are fully aligned with the presented theoretical and empirical results.

\item {\bf Limitations}
    \item[] Question: Does the paper discuss the limitations of the work performed by the authors?
    \item[] Answer: \answerYes{}
    \item[] Justification: The paper discusses limitations in Appendix~\ref{app:limitations}.

\item {\bf Theory assumptions and proofs}
    \item[] Question: For each theoretical result, does the paper provide the full set of assumptions and a complete (and correct) proof?
    \item[] Answer: \answerNA{}
    \item[] Justification: The paper does not include formal theoretical results, theorems, or proofs. Our work is primarily methodological and experimental; all mathematical formulations are used to describe the algorithm and its components, but no formal theorems are claimed or proved. Therefore, this item is not applicable.

    \item {\bf Experimental result reproducibility}
    \item[] Question: Does the paper fully disclose all the information needed to reproduce the main experimental results of the paper to the extent that it affects the main claims and/or conclusions of the paper (regardless of whether the code and data are provided or not)?
    \item[] Answer: \answerYes{}
    \item[] Justification: The paper provides comprehensive details required for reproducibility, including descriptions of all datasets used (see Section “Details of Datasets” and Appendix~\ref{app:details}), implementation details of the proposed algorithm (see “Method” and “Algorithm Overview”), hyperparameter choices, prompt templates (Appendix “Prompt”), and evaluation protocols for each experiment. We also specify the object detection models and baselines used, and state that the code will be publicly released. This level of detail allows other researchers to replicate the main experiments and validate our claims.

\item {\bf Open access to data and code}
    \item[] Question: Does the paper provide open access to the data and code, with sufficient instructions to faithfully reproduce the main experimental results, as described in supplemental material?
    \item[] Answer: \answerYes{}
    \item[] Justification: We state in the abstract and main text that the code will be publicly released. All datasets used in our experiments are from public benchmarks (\lvb, \videomme, \hlvb, \textsc{Ego4D}), and details for data access are provided in Appendix~\ref{app:details}. Instructions for running our framework, data preparation, and experiment replication will be included in the released code repository. Thus, researchers will be able to access both code and data with clear instructions for full reproducibility.

\item {\bf Experimental setting/details}
    \item[] Question: Does the paper specify all the training and test details (e.g., data splits, hyperparameters, how they were chosen, type of optimizer, etc.) necessary to understand the results?
    \item[] Answer: \answerYes{}
    \item[] Justification: The paper specifies all relevant experimental details, including descriptions of dataset splits, hyperparameters, evaluation metrics, and prompt templates (see “Experiment,” Table captions, and Appendix~\ref{app:details}). As our method is training-free, we clarify in the main text which components rely on pre-trained models and explicitly describe all parameter settings for reproducibility. This ensures that readers can fully understand and interpret the reported results.

\item {\bf Experiment statistical significance}
    \item[] Question: Does the paper report error bars suitably and correctly defined or other appropriate information about the statistical significance of the experiments?
    \item[] Answer: \answerNo{}
    \item[] Justification: The paper does not report error bars or formal statistical significance tests for the main experimental results, as our approach is deterministic and uses fixed dataset splits and pre-trained models. Metrics are reported as single values following common practice in recent long video QA benchmarks. While this is standard in the area, we acknowledge that including error bars or additional significance analysis would further strengthen the experimental evaluation.

\item {\bf Experiments compute resources}
    \item[] Question: For each experiment, does the paper provide sufficient information on the computer resources (type of compute workers, memory, time of execution) needed to reproduce the experiments?
    \item[] Answer: \answerYes{}
    \item[] Justification: The paper specifies the computing environment in Appendix~\ref{app:system_specifications}, and 
    reports both latency and FLOPs for major baselines and our method in Table~\ref{tab:main_bench_efficiency}. We also provide the number of iterations, average processing time, and model sizes in the main text and tables. This information is sufficient for others to estimate compute requirements and reproduce the experiments.
    
\item {\bf Code of ethics}
    \item[] Question: Does the research conducted in the paper conform, in every respect, with the NeurIPS Code of Ethics \url{https://neurips.cc/public/EthicsGuidelines}?
    \item[] Answer: \answerYes{}
    \item[] Justification: The research follows the NeurIPS Code of Ethics. All datasets used are publicly available, appropriately licensed, and include human annotation with proper privacy safeguards (see Appendix~\ref{app:details}). No personally identifiable information or sensitive data is used. The proposed methods and experiments present no foreseeable risk of harm, discrimination, or privacy violation. Anonymity is preserved in all supplementary materials.

\item {\bf Broader impacts}
    \item[] Question: Does the paper discuss both potential positive societal impacts and negative societal impacts of the work performed?
    \item[] Answer: \answerYes{}
    \item[] Justification: Our paper discusses broader impacts in Appendix~\ref{app:broader_impacts}.
    
\item {\bf Safeguards}
    \item[] Question: Does the paper describe safeguards that have been put in place for responsible release of data or models that have a high risk for misuse (e.g., pretrained language models, image generators, or scraped datasets)?
    \item[] Answer: \answerNA{}
    \item[] Justification: Our work introduces a semantic-logical search framework for keyframe selection that builds upon existing object detection models and benchmarks. It does not release new datasets scraped from the internet or high-risk generative models. While our method improves video understanding capabilities, it doesn't introduce fundamentally new capabilities that would require specific safeguards beyond those already in place for the underlying technologies (such as YOLO-World) that we utilize.
    \begin{itemize}
        \item The answer NA means that the paper poses no such risks.
        \item Released models that have a high risk for misuse or dual-use should be released with necessary safeguards to allow for controlled use of the model, for example by requiring that users adhere to usage guidelines or restrictions to access the model or implementing safety filters. 
        \item Datasets that have been scraped from the Internet could pose safety risks. The authors should describe how they avoided releasing unsafe images.
        \item We recognize that providing effective safeguards is challenging, and many papers do not require this, but we encourage authors to take this into account and make a best faith effort.
    \end{itemize}

\item {\bf Licenses for existing assets}
    \item[] Question: Are the creators or original owners of assets (e.g., code, data, models), used in the paper, properly credited and are the license and terms of use explicitly mentioned and properly respected?
    \item[] Answer: \answerNA{}
    \item[] Justification: Our work introduces a semantic-logical search framework for keyframe selection that builds upon existing object detection models and benchmarks. It does not release new datasets scraped from the internet or high-risk generative models. While our method improves video understanding capabilities, it doesn't introduce fundamentally new capabilities that would require specific safeguards beyond those already in place for the underlying technologies (such as YOLO-World) that we utilize.

\item {\bf New assets}
    \item[] Question: Are new assets introduced in the paper well documented and is the documentation provided alongside the assets?
    \item[] Answer: \answerYes{}
    \item[] Justification: We will release code for our \fancynameshort framework upon publication, as mentioned in the abstract. The code will be accompanied by comprehensive documentation detailing the implementation of our four logical dependencies (spatial, temporal, attribute, and causal), the iterative refinement process, and instructions for reproducing our experimental results. Our paper does not introduce new datasets but rather evaluates our method on existing benchmarks including \lvb, \videomme, and \hlvb, which are properly cited throughout the paper.

\item {\bf Crowdsourcing and research with human subjects}
    \item[] Question: For crowdsourcing experiments and research with human subjects, does the paper include the full text of instructions given to participants and screenshots, if applicable, as well as details about compensation (if any)? 
    \item[] Answer: \answerNA{}
    \item[] Justification: Our research does not involve crowdsourcing or human subject experiments. We evaluate our method using existing benchmarks (\lvb, \videomme, \lvb) that contain human-annotated ground truth data, but we did not collect new human annotations or conduct human evaluations as part of our work. Our methodology is purely algorithmic, focusing on the semantic-logical frameworks for keyframe selection and evaluation through computational metrics.

\item {\bf Institutional review board (IRB) approvals or equivalent for research with human subjects}
    \item[] Question: Does the paper describe potential risks incurred by study participants, whether such risks were disclosed to the subjects, and whether Institutional Review Board (IRB) approvals (or an equivalent approval/review based on the requirements of your country or institution) were obtained?
    \item[] Answer: \answerNA{}
    \item[] Justification: Our research does not involve human subjects. We utilize existing benchmark datasets (\lvb, \videomme, \hlvb) without collecting new data from human participants. Our work focuses on developing and evaluating algorithmic approaches for keyframe selection based on semantic-logical relationships, which do not require IRB approval or equivalent ethical review processes.

\item {\bf Declaration of LLM usage}
    \item[] Question: Does the paper describe the usage of LLMs if it is an important, original, or non-standard component of the core methods in this research? Note that if the LLM is used only for writing, editing, or formatting purposes and does not impact the core methodology, scientific rigorousness, or originality of the research, declaration is not required.
    %this research? 
    \item[] Answer: \answerYes{}
    \item[] Justification: Our \fancyname framework uses LLMs (specifically mentioned in Section~\ref{sec:evaluation_metrics} and Figure~\ref{fig:framework}) as part of our query decomposition process. We employ models such as \textsc{LLaVA-7B} and \textsc{GPT-4o} to extract semantic information from textual queries, including key objects, cue objects, and their logical relationships. This LLM-based decomposition is an integral component of our method, as it enables the identification of the four logical relation types (spatial, temporal, attribute, and causal) that guide our keyframe selection process. The prompt template for this query grounding is provided in Appendix~\ref{app:prompt}.

\end{enumerate}

\end{document}

%% file: sec/abstract.tex
\vspace{-1em}
\begin{abstract}
\vspace{-0.5em}
Understanding long video content is a complex endeavor that often relies on densely sampled frame captions or end-to-end feature selectors, yet these techniques commonly overlook the logical relationships between textual queries and visual elements. In practice, computational constraints necessitate coarse frame subsampling, a challenge analogous to ``finding a needle in a haystack.'' To address this issue, we introduce a semantics-driven search framework that reformulates keyframe selection under the paradigm of \fancyname. Specifically, we systematically define four fundamental logical dependencies: 1) spatial co-occurrence, 2) temporal proximity, 3) attribute dependency, and 4) causal order. These relations dynamically update frame sampling distributions through an iterative refinement process, enabling context-aware identification of semantically critical frames tailored to specific query requirements. Our method establishes new SOTA performance on the manually annotated benchmark in key-frame selection metrics. Furthermore, when applied to downstream video question-answering tasks, the proposed approach demonstrates the best performance gains over existing methods on \lvb and \videomme, validating its effectiveness in bridging the logical gap between textual queries and visual-temporal reasoning. The code will be publicly available.
\end{abstract}
\vspace{-0.5em}

%% file: sec/intro.tex
\vspace{-1.25em}
\section{Introduction}
\vspace{-0.5em}

Vision-Language Models (VLMs)~\cite{yin2024survey} have achieved remarkable progress in video understanding~\cite{zou2024seconds, tang2023video}, particularly in video question answering~\cite{wang2024gpt4video, 84}, demonstrating potential for modeling real-world scenarios. However, existing methods can only simultaneously process a limited number of frames due to the inherent token limit and extremely high dimension of spatio-temporal video data, especially for long videos. Furthermore, uniformly sampled keyframes are query-agnostic and insufficient to represent query-related contents. To tackle these challenges, this paper addresses a pivotal research question:

\vspace{-0.5em}
\begin{quote}
\textit{How can we efficiently and accurately select keyframes that are semantically critical for answering video-based queries?}
\end{quote}
\vspace{-0.5em}

We hypothesize that deconstructing visual semantic and logical cues (e.g., target objects, logical relations including \textit{temporal}, \textit{spatial}, \textit{attribute}, and \textit{causal} relationships between visual entities) from textual queries enables effective identification of task-relevant frames through heuristic sampling and search. Building on this insight, we propose \textbf{\fancyname (\fancynameshort)}, a novel keyframe search framework that incorporates target object confidence estimation and joint verification of visual semantic logic into the iterative update of frame sampling distribution and selects the most informative frames with the highest confidence. Experimental results show that our approach requires only sparse sampling (1.4\% of frames per video on average) to identify critical frames, significantly reducing computational complexity compared to conventional dense sampling strategies while maintaining performance on downstream video understanding tasks.

\begin{figure}[t]
    \centering
    \includegraphics[width=\textwidth]{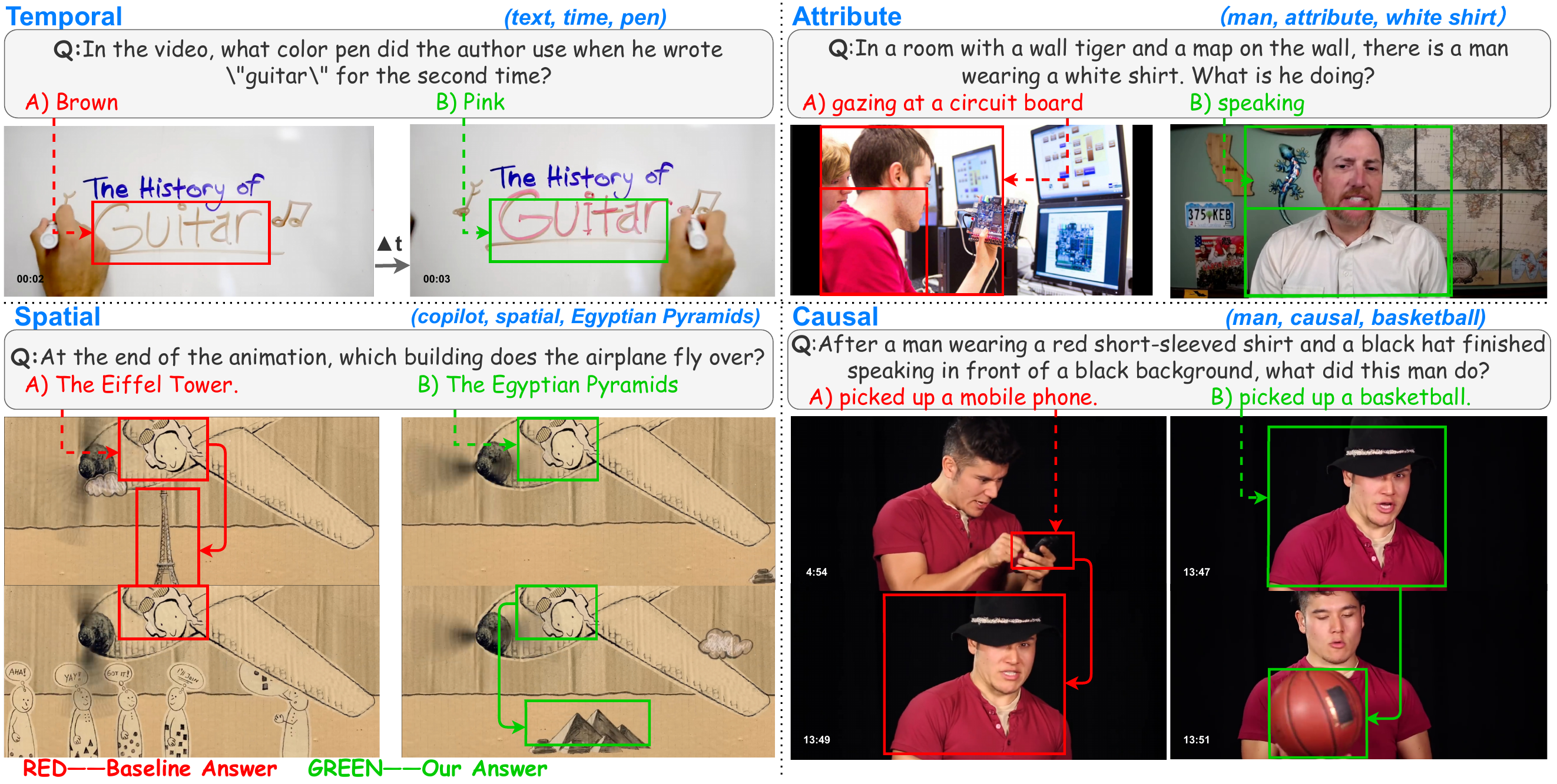}
    \vspace{-1.25em}
    \caption{Examples of four types of visual semantic-logical relationships in video QA detected by our \fancynameshort framework: \textbf{Temporal} \texttt{(text, time, pen)}, \textbf{Attribute} \texttt{(man, attribute, white shirt)}, \textbf{Spatial} (copilot, spatial, Egyptian Pyramids), and \textbf{Causal} \texttt{(man, causal, basketball)}. Green boxes indicate correct answers, while red boxes show baseline errors.}
    \vspace{-1.5em}
    \label{fig:first}
\end{figure}

Compared to conventional methods, \fancynameshort shows three distinct advantages. First, the framework is training-free and highly efficient in comparison with dense captioning~\cite{chen2024sharegpt4video,kim2024you,wang2024videoagent} or video clustering~\cite{wang2024videotree,rajan2025key} strategies, sampling only 1.4\% of frames on average in \textsc{LVHaystack}. Second, it explicitly models logical binary relations (namely spatial, temporal, attribute, and causal) in the query beyond simple target detection~\cite{ye2025rethinking}, utilizing additional visual semantic features and enhancing logical consistency throughout the reasoning process. Third, \fancynameshort is a plug-and-play module, which can be seamlessly integrated into existing VLM pipelines without cross-component dependencies. 

We further examine \fancynameshort on several public datasets, including \lvb~\cite{lvhdataset}, a comprehensive benchmark for long video understanding; \videomme~\cite{videomme}, a widely adopted multimodal video question answering dataset; and \hlvb~\cite{lvhdataset} with meticulously annotated keyframes based on human feedback for more precise analysis. Extensive experiments demonstrate significant improvements in both the semantic similarity and temporal coverage between the retrieved keyframes and the ground truth labels, as well as the accuracy in downstream video question-answering tasks. More importantly, with only \textbf{1.4\%} of video frames (\textsc{Ego4D}~\cite{grauman2022ego4d}) sampled in the search iteration, our method achieves an \textbf{8.7\%} improvement in \textsc{GPT-4o}~\cite{hurst2024gpt}'s long video QA accuracy. This performance gain is attributed to our simple yet powerful observation: query-guided visual semantic logic retrieval can mitigate the gap between potential visual logic in video frames and the logic expressed in the query. To be specific, constructing ternary logic triplets with visual elements (e.g., \texttt{object1, logic type, object2}) can enhance downstream reasoning capabilities when performing textual-visual retrieval.

To the best of our knowledge, we are arguably the first to search for keyframes in long videos by detecting visual semantic logic, with potential extensions to other textual-visual retrieval tasks. Our main contributions are as follows:
\begin{itemize}
    \item We define four fundamental types of semantic logic relations in video QA tasks, including \textit{temporal}, \textit{causal}, \textit{attribute}, and \textit{spatial} relations, which can be accurately detected across various datasets.
    \item We sample only 1.4\% of frames on average of frames on average during keyframe search through heuristic sampling and distribution updating by different visual semantics and logical relations.
    \item We comprehensively evaluate retrieval efficiency, semantic similarity, temporal coverage, and video question answering accuracy across several widely used video understanding datasets, demonstrating significant improvements in downstream tasks.
\end{itemize}

%\clearpage

%% file: sec/Method.tex
\vspace{-1.em}
\section{Method}
\vspace{-0.75em}

\begin{figure*}[t]
\centering
\includegraphics[width=\textwidth]{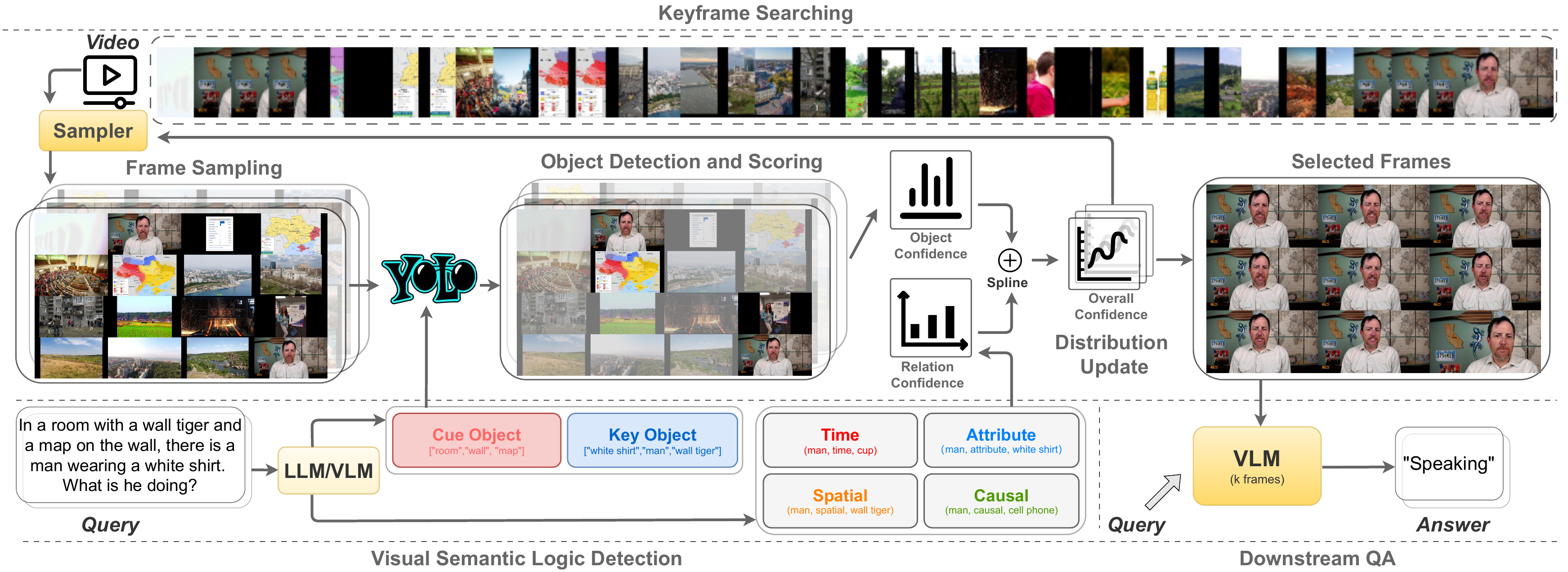}
\vspace{-1.75em}
\caption{\textbf{Our \fancynameshort Framework for Efficient Keyframe Selection.}
\fancynameshort sparsely samples frames and selects key ones via object detection and logic verification. Steps: 1) Use LLM\&VLM to extract cue/target objects and four logic types (\textit{spatial}, \textit{temporal}, \textit{attribute}, \textit{causal}); 2) Adaptive sampling with evolving confidence; 3) Detect objects via \textsc{YOLO-World}; 4) Fuse scores with a spline function to identify high-confidence frames for downstream tasks.}
\vspace{-1.5 em}
\label{fig:framework}
\end{figure*}

Although existing long-context VLM frameworks implement keyframe search for video QA tasks~\cite{35,47,57,65,68,80}, their computational efficiency and searching accuracy remain suboptimal. To address this \textit{needle-in-a-haystack} challenge~\cite{61,zhao2024needle}, we propose a novel method \fancynameshort that aligns the semantic relations between the text modality and video modality, enhancing the plausibility of logical reasoning and performance of downstream tasks.

\vspace{-0.5em}
\subsection{Task Formulation}
\vspace{-0.5em}
Given a video sequence $V = \{f_t\}_{t=1}^{N_{v}}$ with $N_{v}$ frames and a query $Q$, the ideal temporal search framework aims to retrieve the minimal keyframe subset $V^K = \{f_{m_i}\}_{i=1}^K \subseteq V$ with $K$ keyframes that satisfies:
\begin{itemize}
    \item \textbf{Conservation}: The keyframe subset $V^K  \subseteq V$ must satisfy the answer consistency condition: $\mathcal{A}(V^K, Q) = \mathcal{A}(V, Q)$, where $\mathcal{A}(\cdot)$ denotes the video QA function.
    \item \textbf{Compactness}: $V^K$ must be a minimal subset that preserves completeness, which means that no frame in $V^K$ can be removed without hindering the accuracy and efficiency of video QA.
\end{itemize}

\vspace{-0.5em}
\subsection{Visual Semantic Logic Extraction}
\vspace{-0.5em}
\label{subsec:semantic_logic}

Starting from a question \( Q \) and uniformly sampled frames \( \overline{V}_N \) from video \( V \), our goal is to extract key visual elements to answer \( Q \). We first classify the detected objects in \( Q \) and \( \overline{V}_N \) into two categories:
\begin{itemize}
    \item \textbf{Key Objects}: The main participants or references in the scene that the question explicitly or implicitly focuses on (e.g., ``\textit{person}'', ``\textit{microphone}'').
    \item \textbf{Cue Objects}: Secondary or contextual entities that help locate or disambiguate the Key Objects (e.g., ``\textit{book}'', ``\textit{tiger painting}'').
\end{itemize}

\noindent To further leverage semantic and logical links among these objects, we define a set of relations 
$\mathcal{R} \subseteq \mathcal{O} \times \Delta \times \mathcal{O}$, where each relation $r=(o_i, \delta, o_j) \in \mathcal{R}$, with $o_i, o_j \in \mathcal{O}$ denoting detected objects in the key and cue objects dataset, and $\delta \in \Delta$ representing one of the following types of relations:
\setlength{\emergencystretch}{1em} 

\begin{table}[h!]
\centering
\vspace{-0.75em}
\renewcommand{\arraystretch}{1.1}
\setlength{\tabcolsep}{4pt}
\begin{tabularx}{1.0\textwidth}{>{\hsize=0.23\hsize\bfseries}X >{\hsize=0.77\hsize}X >{\hsize=0.23\hsize\bfseries}X >{\hsize=0.77\hsize}X}
\rowcolor{rowgray}
\multicolumn{2}{c}{\textbf{Spatial Co-occurrence}} & \multicolumn{2}{c}{\textbf{Attribute Dependency}} \\
\multicolumn{2}{X}{
$o_i$ and $o_j$ appear in the same frame, indicating co-occurrence or proximity.
\newline
\textbf{\textit{Example:}} “A person is standing beside a vase.” $\Rightarrow$ \texttt{(person, spatial, vase)}
}
&
\multicolumn{2}{X}{
$o_i$ and $o_j$ share visual properties, e.g., color or size.
\newline
\textbf{\textit{Example:}} “A person wears a black shirt.” $\Rightarrow$ \texttt{(person, attribute, black shirt)}
} \\
\addlinespace[0.5em]
\rowcolor{rowgray}
\multicolumn{2}{c}{\textbf{Temporal Proximity}} & \multicolumn{2}{c}{\textbf{Causal Order}} \\
\multicolumn{2}{X}{
$o_i$ and $o_j$ occur in close frames, linking sequences or transitions.
\newline
\textbf{\textit{Example:}} “After a dog entered the room, a cat entered.” $\Rightarrow$ \texttt{(dog, temporal, cat)}
}
&
\multicolumn{2}{X}{
$o_i$ and $o_j$ follow a cause-effect or prerequisite order.
\newline
\textbf{\textit{Example:}} “A little girl broke the vase.” $\Rightarrow$ \texttt{(little girl, causal, pieces)}
} \\
\end{tabularx}
\end{table}
\vspace{-0.5em}

The choice of these four relations draws on core concepts in linguistics and logic~\cite{cohen1968universals,sowa2000knowledge,talmy2000toward}, which identify spatial, temporal, attributive, and causal aspects as fundamental for structuring, perceiving, and communicating information about events and states. For more details on this selection, please see appendix~\ref{Theoretical} for reference. As shown in Figure~\ref{fig:first}, we construct semantic-logical relations that support a broad range of question-answering tasks. Specifically, questions involving temporal queries (\textit{when does X happen?}''), causal reasoning (\textit{why did Y occur?}’’), attribute dependence (\textit{What is the person wearing sunglasses doing?}''), or spatial constraints (\textit{Who is standing next to the red car?}’’) can be answered more reliably by incorporating these structured relations and contextual cues.

\vspace{-0.5em}
\subsection{Iterative Semantic-Logical Temporal Search}
\vspace{-0.5em}
Based on the extracted key and cue objects and their logic relations, our algorithm iteratively searches for keyframes through semantic and logical reasoning, including four main stages: \textbf{Frame Sampling} (Sec.~\ref{sec:frame_sampling}), \textbf{Object Detection and Scoring} (Sec.~\ref{sec:object_detection_and_scoring}), \textbf{Visual Semantic Logic Detection} (Sec.~\ref{sec:visual_semantic_logic_detection}), and \textbf{Distribution Update} (Sec.~\ref{sec:distribution_update}). The pseudocode is shown in Algorithm~\ref{alg:main}, and Algorithm~\ref{alg:slt_search} provides a more detailed version.
%\vspace{-1.75em}
\begin{algorithm}[t]
\caption{\footnotesize \fancyname\label{alg:main}}
\footnotesize
\SetAlgoNlRelativeSize{0}
\SetNlSty{}{}{:}
\SetKwFunction{SLTSearch}{SemanticLogicalTemporalSearch}
\SetKwFunction{DetectObjects}{DetectObjects}
\SetKwFunction{CalcBaseScore}{CalculateBaseScore}
\SetKwFunction{UpdateScores}{UpdateScores}
\SetKwFunction{CheckSpatial}{CheckSpatialrelation}
\SetKwFunction{CheckTemporal}{CheckTemporalrelation}
\SetKwFunction{CheckCausal}{CheckCausalrelation}
\SetKwFunction{CheckAttribute}{CheckAttributerelation}
\SetKwFunction{DiffuseScores}{DiffuseScores}
\SetKwFunction {ProcessRel}{Processrelation}
\SetKwFunction{Normalize}{NormalizeDistribution}
\SetKwInOut{Input}{Input}\SetKwInOut{Output}{Output}
\SetKwProg{Fn}{Function}{}{}

\Fn{\SLTSearch{$V, Q, K, \Delta_t, \tau, \alpha, \gamma$}}{
    $\mathcal{O}, \mathcal{R} \gets \text{ParseQuestion}(Q)$ \hfill 
    \textcolor{commentgray}{\scriptsize // Extract key/cue objects and relations}
    
    $P \gets \text{Uniform}, B \gets |V|, S \gets \emptyset, N_v \gets |V|$ \hfill 
    \textcolor{commentgray}{\scriptsize // Initialize distribution and state}
    
    \While{$B > 0$ \textbf{and} $|\mathcal{O}| > 0$}{
        $k \gets \lfloor\sqrt{B}\rfloor$, $G \gets \Grid(\Sample(P, k^2))$ \hfill 
        \textcolor{commentgray}{\scriptsize // Adaptive grid sampling}
        
        $\Omega \gets \DetectObjects(G)$ \hfill 
        \textcolor{commentgray}{\scriptsize // Detect objects in sampled frames}
        
        \ForEach{$t \in G$}{
            $C_t \gets \CalcBaseScore(\Omega_t)$ \hfill 
            \textcolor{commentgray}{\scriptsize // Base detection confidence}
            
            \ForEach { $r_{type} \in \mathcal {R}$ } { 
                $\delta \gets \ProcessRel (r_{type}, \Omega , \Delta _t, \tau , \alpha , \gamma )$ \hfill 
                \textcolor{commentgray}{\scriptsize // relations require distinct processing}
                
                $C_t \gets C_t + \delta $ 
            } 
            $\UpdateScores(S, t, C_t)$ \hfill 
            \textcolor{commentgray}{\scriptsize // Update global score registry}
        }
        
        $\DiffuseScores(S, w)$ \hfill 
        \textcolor{commentgray}{\scriptsize // Temporal context propagation}
        
        $P \gets \Normalize(S)$, $B \gets B - k^2$ \hfill 
        \textcolor{commentgray}{\scriptsize // Update sampling distribution}
        
        \ForEach{$g \in \TopK(S, K)$}{
            \If{$\Omega[g] \cap \mathcal{O} \neq \emptyset$}{
                $\mathcal{O} \gets \mathcal{O} \setminus \Omega[g]$ \hfill \textcolor{commentgray}{\scriptsize // Remove identified key objects}
            }
        }
    }
    \Return $\TopK(S, K)$ \hfill 
    \textcolor{commentgray}{\scriptsize // Return top-K keyframes}
}
\end{algorithm}
\vspace{-0.5em}
\subsubsection{Frame Sampling}
\label{sec:frame_sampling}
\vspace{-0.5em}
To accelerate the search process, we avoid exhaustively scanning all $N_v$ video frames and instead employ a distributed sampling strategy. Let $N_v$ denote the total number of frames in the video, and $P$ be a uniformly initialized sampling distribution over all frames. The sampling process is then defined as:
\begin{equation}
    I_s = \mathrm{Sample}(P \odot N_v, N_s),
\end{equation}
where $\mathrm{Sample}(\cdot, N_s)$ selects a subset of $N_s$ frames according to the distribution $P \odot N_v$.  To further leverage the detecting ability of YOLO, we stack the sampled frames into a $k \times k$ grid, which imposes a constraint on the sample size $N_s$. Specifically, we require:
\begin{equation}
    N_s \in \{k^2 \mid k \in \mathbb{Z}\}
    \quad \text{and} \quad
    N_s < N_v.
\end{equation}
In practice, this ensures that the number of sampled frames can be reshaped into a compact 2D grid for efficient processing. Although $P$ is initially uniform, it can be adapted over multiple rounds of sampling to focus on frames of higher interest in the video.
\vspace{-0.5em}
\subsubsection{Object Detection and Scoring}
\label{sec:object_detection_and_scoring}
\vspace{-0.5em}

In this stage, we construct the detection search space by taking the union of both key objects and cue objects. For each iteration, we detect objects on the $N_s$ sampled frames using a lightweight model like \textsc{YOLO-World}~\cite{yolo} for high efficiency and score the frames based on detection confidence. Specifically, let $\Omega_{t}$ be the set of detected objects in the frame at time $t$, $c_{o}$ the confidence of each detected object, and $w_{o}$ the corresponding weight. We define the frame score as:
\begin{equation}
    C_{t} = \max_{o \in \Omega_{t}} \bigl(c_{o} \cdot w_{o}\bigr).
\end{equation}
If the confidence score of any key object exceeds a predefined threshold, it is added to a list, thereby maintaining a record of frames where crucial targets have been identified for subsequent processing.

\vspace{-0.5em}
\subsubsection{Visual Semantic Logic Detection}
\label{sec:visual_semantic_logic_detection}
\vspace{-0.5em}
\begin{figure*}[t]
    \centering
    \includegraphics[width=\linewidth]{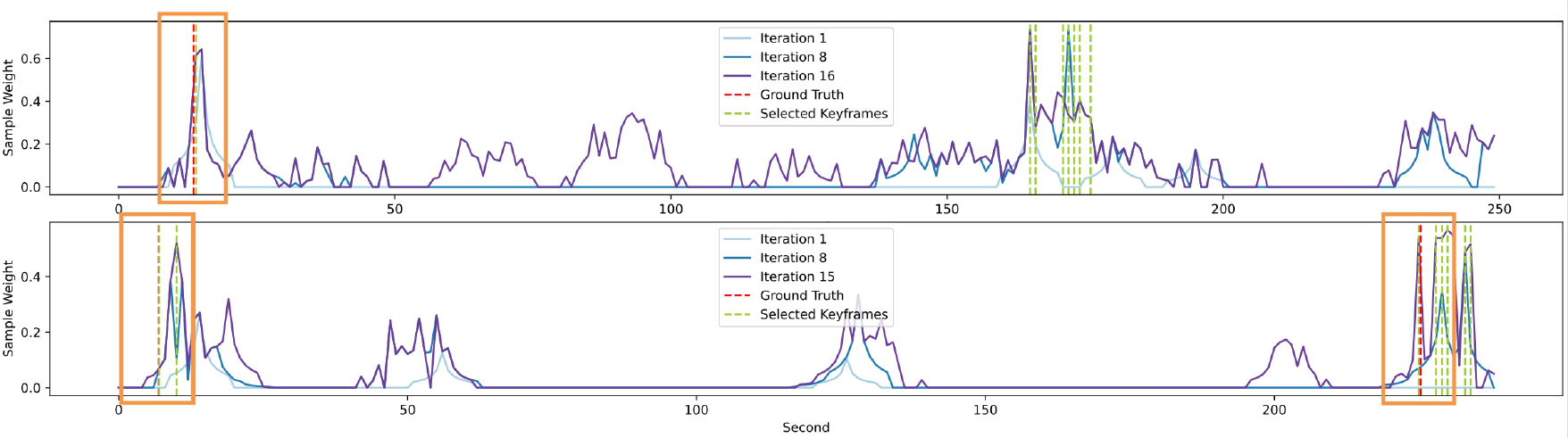}
    \caption{\textbf{Sample weight evolution under \fancynameshort optimization for keyframe selection.}
        Top: 16 iterations show progressive convergence toward Ground Truth (red). Bottom: 15 iterations demonstrate similar alignment. Yellow highlights indicate precise matches between algorithm outputs (green) and manual annotations.}
    \label{fig:revolution}
    \vspace{-1.5em}
\end{figure*}

Beyond individual object detection and frame-level scoring, we refine each frame’s confidence score by modeling higher-order object relations. Let $\mathcal{R}$ be the set of relations, where each $r \in \mathcal{R}$ involves a pair $(o_1, o_2)$ and is labeled by a type $r_{\text{type}}$. Denote $C_t$ as the confidence score at time $t$, with a global scaling factor $\alpha$ and a relation-specific weight $\gamma_{r_{\text{type}}}$ controlling each logic type’s impact. The refined confidence $C_t^{(r)}$ after applying relation $r$ is defined as:
\begin{equation}
    C_t^{(r)} = C_t + \alpha \cdot \gamma_{r_{\text{type}}}.
\end{equation}
\vskip -0.5em

\noindent \textbf{Spatial Relation.} A \textit{spatial} relation enforces that two objects $o_1$ and $o_2$ must co-occur in the same frame. Let $\Omega_t$ be the set of detected objects in frame $t$. If both $o_1 \in \Omega_t$ and $o_2 \in \Omega_t$, then the corresponding frame confidence is updated as:
\begin{equation}
    C_t \leftarrow C_t + \alpha \cdot \gamma_{\text{spatial}}.
\end{equation}
\vskip -0.5em

\noindent \textbf{Attribute Relation.} An \textit{attribute} relation is satisfied when $o_1$ and $o_2$ share sufficient bounding-box overlap in the same frame. Let $\mathrm{overlap}$ be the ratio of their intersection area to the minimum of their individual bounding-box areas. If the overlap ratio exceeds a predefined threshold $\tau$ ($\tau=0.5$ in our experimental setting), we increase the frame confidence:
\begin{equation}
    C_t \leftarrow C_t + \alpha \cdot \gamma_{\text{attribute}}.
\end{equation}
\vskip -0.5em

\noindent \textbf{Time Relation.} A \textit{time} relation checks whether two objects appear in temporally close frames. Suppose $t_i$ and $t_j$ ($t_i \le t_j$) are sampled such that $|\,t_j - t_i| < \Delta_t$, where $\Delta_t$ is a  threshold (e.g.\ 5 frames in our experimental setting), if $o_1$ occurs in frame $t_i$ and $o_2$ in frame $t_j$, then both frames' confidences are updated:
\begin{equation}
    C_{t_i} \leftarrow C_{t_i} + \alpha \cdot \gamma_{\text{time}}, 
    \quad
    C_{t_j} \leftarrow C_{t_j} + \alpha \cdot \gamma_{\text{time}}.
\end{equation}
\vskip -0.5em

\noindent \textbf{Causal Relation.} A \textit{causal} relation models an ordering constraint, enforcing that $o_1$ must appear at an earlier time than $o_2$. Specifically, if $o_1 \in \Omega_{t_i}$ and $o_2 \in \Omega_{t_j}$ with $t_i < t_j$, we update the confidences of frames $t_i$ and $t_j$ by:
\begin{equation}
    C_{t_i} \leftarrow C_{t_i} + \alpha \cdot \gamma_{\text{causal}}, 
    \quad
    C_{t_j} \leftarrow C_{t_j} + \alpha \cdot \gamma_{\text{causal}}.
\end{equation}
\vskip -0.5em

Through this scoring mechanism, frames with detected relations will have greater confidence and are more likely to be retrieved as keyframes for the given query and video. We have also conducted hyperparameter search experiments, and find that $\alpha=0.3$ (from 0.3, 0.5, 0.7, 1.0) and $\gamma_{r_{\text{type}}}=0.5$ achieve the best results across different datasets. 

\vspace{-1.0em}
\subsubsection{Distribution Update}
\label{sec:distribution_update}
\vspace{-0.5em}

After each iteration of frame sampling, we merge newly obtained frame confidences into the global score distribution $\{S_f\}$ spanning all frames $f = 1, 2, \ldots, N_v$. When a frame $f$ is selected for detection, its score is assigned to the confidence value $C_f$, and the visitation counter $N_{v,f}$ is reset to 0.
To incorporate temporal context, we diffuse this updated score to neighboring frames within a window of size $w$. Denoting each nearby index by $f \pm \delta$ (for $\delta \in [-w, w]$), we apply:
\begin{equation}
    S_{f \pm \delta} \leftarrow
    \max\left(
        S_{f \pm \delta}, \,
        \frac{S_f}{1 + |\delta|}
    \right).
    \label{eq:dist_update_window}
\end{equation}
In this way, high-confidence frames raise the scores of close-by frames, reflecting temporal continuity. Following these local updates, the sampling distribution $P$ is refined using spline interpolation, and then normalized. This iteration proceeds until either the search budget $B$ is reached or all key objects have been successfully identified. The visualization of distribution in different iterations can be seen in Figure~\ref{fig:revolution}.  Finally, the method outputs the top $K$ frames according to their terminal scores.

%% file: sec/Experiment.tex
\vspace{-1em}
\section{Experiment}
\vspace{-0.75em}
\subsection{Benchmark Datasets}
\vspace{-0.5em}
The proposed \fancynameshort is systematically evaluated across four benchmark datasets: a) \lvb~\cite{lvhdataset} for assessing long-context video-language comprehension capabilities; b) \videomme~\cite{videomme} as the first comprehensive benchmark for multimodal video analytics; c) \hlvb, extended from \lvb with human-annotated frame index answers; and d) \textsc{Haystack-Ego4D}, derived from \textsc{Ego4D} with similar annotations. While \lvb and \videomme measure performance enhancement in QA accuracy, \textsc{Haystack-Ego4D} and \hlvb quantitatively evaluate keyframe selection accuracy through recall and precision metrics. Further details of datasets are provided in Appendix~\ref{app:details}.

\vspace{-0.75em}
\subsection{Evaluation Metrics}
\label{sec:evaluation_metrics}
\vspace{-0.5em}

\subsubsection{Evaluation Metrics for Search Utility}
\vspace{-0.5em}
Our assessment framework emphasizes both effectiveness and efficiency. For search effectiveness, we use three metrics to compare model-predicted keyframes with human annotations, considering both individual frames and full sets—addressing the possibility of multiple valid keyframe sets per query. For frame-level comparison, we evaluate the alignment between a predicted frame $f_\text{pt}$ and a human-annotated frame $f_\text{gt}$ from two perspectives:
\vspace{-0.25em}

\noindent \textbf{Temporal coverage} evaluates the coverage of ground truth frames by predicted frames in the temporal perspective, which can be described as:
\vspace{-0.25em}
\begin{equation}
    T_{\text{cover}}(T_{\text{pt}}, T_{\text{gt}}) = 
    \frac{\sum\limits_{i=1}^{|N_{\text{gt}}|} 
    \mathbb{I} \left[ \min\limits_{j} \left| t_{\text{gt}}^i - t_{\text{pt}}^j \right| \leq \delta \right]}
    {|N_{\text{gt}}|},
\end{equation}

where $T_\text{pt}$ and $T_\text{gt}$ denote the sets of predicted and ground truth timestamps, respectively. Here, $|N_\text{gt}|$ is the number of ground truth frames, $t_\text{gt}^i$ and $t_\text{pt}^j$ are the $i$-th ground truth and $j$-th predicted timestamps, respectively. $\delta$ is the temporal similarity threshold defining the maximum allowed time deviation, and $\mathbb{I}[\cdot]$ is the indicator function, returning 1 if the condition holds and 0 otherwise.

\vspace{-0.25em}
\noindent \textbf{Visual Similarity} is measured by the Structural Similarity Index (SSIM)~\cite{ssim}, capturing structural detail, luminance, and contrast between $f_\text{pt}$ and $f_\text{gt}$.
For set-to-set comparison, the key challenge is defining inter-set similarity. We adopt \textbf{Precision} $P$ and \textbf{Recall} $R$ as complementary metrics: Precision checks whether each predicted frame matches any reference frame, while Recall ensures that all reference frames are represented.
Given the ground truth set $F_{\text{gt}} ={f^{j}{\text{gt}}}^n{j=1}$ and the predicted set $F_{\text{pt}} ={f^{i}{\text{pt}}}^m{i=1}$, we define the multimodal retrieval quality metrics as follows:
\vskip -0.5em
\begin{subequations}
\begin{empheq}[left=\empheqlbrace]{align}
P(F_{\text{pt}}, F_{\text{gt}}) &= \frac{1}{|F_{\text{pt}}|} \sum_{f^{i}_{\text{pt}} \in F_{\text{pt}}} \max_{f^{j}_{\text{gt}} \in F_{\text{gt}}} \phi(f^{i}_{\text{pt}}, f^{j}_{\text{gt}}),\label{eq:precision} \\
R(F_{\text{pt}}, F_{\text{gt}}) &= \frac{1}{|F_{\text{gt}}|} \sum_{f^{j}_{\text{gt}} \in F_{\text{gt}}} \max_{f^{i}_{\text{pt}} \in F_{\text{pt}}} \phi(f^{j}_{\text{gt}}, f^{i}_{\text{pt}}), \label{eq:recall}
\end{empheq}
\end{subequations}
\vspace{-0.5em}
where $\phi(\cdot,\cdot)$ represents an extensible multimodal similarity metric function. 

\vspace{-0.5em}
\subsubsection{Evaluation Metrics for Search efficiency}
\vspace{-0.5em}

Existing studies~\cite{13,47,65,68,71} have mainly concentrated on optimizing task-specific performance metrics while neglecting computational efficiency in temporal search operations. To systematically analyze this dimension, our evaluation framework incorporates two criteria: 1) \textbf{FLOPs} representing arithmetic operation complexity, and 2) \textbf{Latency} recording real-world execution duration.

\vspace{-0.5em}
\subsection{Evaluation of Search Framework efficiency}
\vspace{-0.5em}
\begin{table*}[t]
    \centering
    \setlength\tabcolsep{8pt}
    \setlength\extrarowheight{1pt}
    \arrayrulecolor[gray]{0.7}
    \begin{adjustbox}{width=\linewidth}
\begin{tabular}{l|c|c|c|c|c|c|c}
        \toprule
        \multirow{2}{*}{\bf Method} &  \multirow{2}{*}{\bf Training Required} &  \multicolumn{4}{c|}{\bf Searching Efficiency}  &   \multicolumn{2}{c}{\bf Overall Task Efficiency} \\  
        & & Matching & Iteration & TFLOPs $\downarrow$ & Latency (sec) $\downarrow$ & Latency (sec) $\downarrow$ & Acc $\uparrow$ \\
        \midrule
        \multicolumn{8}{c}{Static Frame Sampling} \\ 
        \hline
        \textsc{Uniform-8}~\cite{lvhdataset}& Training-Based & N/A & N/A & N/A & 0.2 & 3.8 & 53.7  \\    
        \midrule
        \multicolumn{8}{c}{Dense Retrieval} \\ 
        \hline
        \textsc{VideoAgent}~\cite{13}& Training-Based & CLIP-1B~\cite{radford2021learning} & 840 & 536.5 & 30.2 & 34.9 & 49.2 \\
        \textsc{T\textasteriskcentered-Retrieval}~\cite{ye2025rethinking} & Training-Based & \textsc{YOLO-World}-110M & 840 & 216.1 & 28.6 & 32.2 & 57.3\\
        \midrule
        \multicolumn{8}{c}{Temporal Search} \\
        \hline
        \textsc{T\textasteriskcentered-Attention}~\cite{ye2025rethinking}& Training-Based & N/A & N/A & 88.9 & 13.7 & 17.3 & 59.3 \\
        \textsc{T\textasteriskcentered-Detector}~\cite{ye2025rethinking}& \textbf{Training-Free} & \textsc{YOLO-World}-110M & 43 & 31.7 & 7.3 & 11.1 & 59.8 \\
        \rowcolor{gray!10}
        \textbf{\textsc{\fancynameshort(ours)-Detector}} & \textbf{Training-Free} & \textsc{YOLO-World}-110M & 49 & 33.3 & 7.8 & 11.6 & \textbf{61.5} \\
        \bottomrule
    \end{tabular}
    \end{adjustbox}
    \caption{Evaluation of performance metrics across the \lvh benchmark, presenting both search efficiency and end-to-end processing overhead (combining search and inference stages).}
    \label{tab:main_bench_efficiency}
    \vspace{-1.5em}
\end{table*}
Current approaches for keyframe selection can be broadly categorized into three paradigms: statistic-based frame sampling, dense feature retrieval-based selection, and temporal search-based methods. As shown in Table \ref{tab:main_bench_efficiency}, while uniform sampling achieves the fastest processing speed, its ignorance of frame semantics severely limits downstream task effectiveness. Although dense feature retrieval methods attain moderate accuracy improvements (57.3\%), their exhaustive frame processing demands 4.2$\times$ more TFLOPs and introduces 4.5$\times$ higher latency than our temporal search approach. Crucially, our method introduces four visual semantic logic detectors during temporal search while maintaining comparable execution time to \tstar methods. This strategic design elevates downstream task accuracy to 61.5\%, achieving the best performance-efficiency trade-off.

\vspace{-0.5em}
\subsection{Visual Semantic Logic Search Performance}
\vspace{-0.5em}

As demonstrated in Table~\ref{tab:framesearch_evaluation}, we evaluate \fancynameshort on \lvb from two critical perspectives: visual similarity (measured by precision and recall) and temporal coverage. Our method achieves state-of-the-art performance across all metrics. Specifically, under the 32-frame setting, \fancynameshort attains a precision of 74.5\% and recall of 92.5\%, outperforming all baselines in visual similarity. More notably, the temporal coverage of \fancynameshort reaches 41.4\%, surpassing the second-best method (\tstar at 36.5\%) by 13.4\%—the largest margin among all comparisons. This significant improvement highlights the effectiveness of our visual semantic logic detection modules in identifying query-relevant keyframes with both semantic alignment and temporal completeness.

\begin{wraptable}{r}{0.48\linewidth}
  \vspace{-1\baselineskip}
  \centering
  \footnotesize            
  \setlength{\tabcolsep}{3pt}
  \setlength\extrarowheight{1pt}

  \caption{Search utility results on \lvb. 
    Best scores in the 8-frame setting are \underline{underlined}, 
    and in the 32-frame setting are \textbf{bold}. 
    \textcolor{gray}{Gray} indicates results from the original paper.}
  \label{tab:framesearch_evaluation}

  \begin{adjustbox}{width=\linewidth}
  \begin{tabular}{lcccc}
  \hline
  \multicolumn{1}{l|}{\multirow{2}{*}{\textbf{Method}}} &
  \multicolumn{1}{c|}{\multirow{2}{*}{\textbf{Frame}}} &
  \multicolumn{3}{c}{\textbf{\lvb}} \\
  \multicolumn{1}{l|}{} & \multicolumn{1}{c|}{} &
  Precision $\uparrow$ & Recall $\uparrow$ & Time $\uparrow$ \\ \hline
  \multicolumn{5}{c}{Static Frame Sampling Method} \\ \hline
  \color{gray}\textsc{Uniform}~\cite{lvhdataset} & \color{gray}8  & \color{gray}56.0 & \color{gray}72.0 & \color{gray}6.3 \\
  \rowcolor{gray!20}
  \textsc{Uniform}                                    & 8  & 60.7 & 80.4 & 4.7 \\
  \color{gray}\textsc{Uniform}                        & \color{gray}32 & \color{gray}58.7 & \color{gray}81.6 & \color{gray}24.9 \\
  \rowcolor{gray!20}
  \textsc{Uniform}                                    & 32 & 60.2 & 85.0 & 8.1 \\ \hline
  \multicolumn{5}{c}{Dense Retrieval Method} \\ \hline
  \color{gray}\textsc{VideoAgent}~\cite{13}           & \color{gray}10.1 & \color{gray}58.8 & \color{gray}73.2 & \color{gray}8.5 \\
  \color{gray}\textsc{Retrieval-based}~\cite{ye2025rethinking} & \color{gray}8 & \color{gray}63.1 & \color{gray}65.5 & \color{gray}6.3 \\
  \color{gray}\textsc{Retrieval-based}                & \color{gray}32 & \color{gray}59.9 & \color{gray}80.8 & \color{gray}21.8 \\ \hline
  \multicolumn{5}{c}{Temporal Searching Method} \\ \hline
  \color{gray}\tstar~\cite{ye2025rethinking}          & \color{gray}8  & \color{gray}58.4 & \color{gray}72.7 & \color{gray}7.1 \\
  \rowcolor{gray!20}
  \tstar                                             & 8  & 75.3 & 88.2 & 26.2  \\
  \rowcolor{gray!20}
  \textbf{\fancynameshort{} (ours)}                   & 8  & \underline{75.6} & \underline{88.6} & \underline{26.3} \\
  \color{gray}\tstar                                 & \color{gray}32 & \color{gray}58.3 & \color{gray}83.2 & \color{gray}28.2 \\
  \rowcolor{gray!20}
  \tstar                                             & 32 & 74.0 & 90.3 & 36.5 \\
  \rowcolor{gray!20}
  \textbf{\fancynameshort{} (ours)}                   & 32 & \textbf{74.5} & \textbf{92.5} & \textbf{41.4} \\ \hline
  \end{tabular}
  \end{adjustbox}
  \vspace{-2\baselineskip}
\end{wraptable}

These results empirically support our core hypothesis: leveraging semantic and logical cues from text queries enables precise detection of relevant video frames. Improvements in visual similarity and temporal coverage confirm that \fancynameshort effectively captures keyframes while preserving temporal coherence through visual-logical alignment.

\vspace{-0.5em}
\subsection{Downstream Video QA Performance}
\vspace{-0.5em}
\begin{table*}[t]
    \centering
    \setlength\tabcolsep{3pt}
    \setlength\extrarowheight{1pt}
    \arrayrulecolor[gray]{0.7}
    \begin{adjustbox}{width=\linewidth}
    \begin{tabular}{l|c|c|c|c||l|c|c|c|c}
        \toprule
        \multicolumn{5}{c||}{\bf \lvb} & \multicolumn{5}{c}{\bf \videomme} \\
        \cmidrule{1-10}
        
        \multirow{3}{*}{\bf Model and Size} & \multirow{3}{*}{\bf Frame} & \multicolumn{3}{c||}{\textbf{Video Length}} & \multirow{3}{*}{\bf Model and Size} & \multirow{3}{*}{\bf Frame} & \multicolumn{3}{c}{\textbf{Video Length}} \\
        
        & & \textbf{Long} & \textbf{Medium} & \textbf{Short} & & & \textbf{Long} & \textbf{Medium} & \textbf{Short} \\
        & & 900-3600s & 180-600s & 15-60s & & & 30-60min & 4-15min & 0-2min \\

        \midrule
        
        \textsc{GPT-4o}~\cite{hurst2024gpt} & 8 & 47.1 & 49.4 & 67.3 & \textsc{GPT-4o} & 8 & 55.2 & 60.2 & \textbf{69.6} \\
        \textsc{GPT-4o} + \tstar & 8 & 49.1 & 56.2 & 68.0 & \textsc{GPT-4o} + \tstar  & 8 & 55.2 & \textbf{61.2} & 68.9 \\
        \rowcolor{gray!10}
        \arrayrulecolor{gray!40}\hline \arrayrulecolor{black}
        \textsc{GPT-4o} + \fancynameshort(ours)  & 8 & \textbf{51.2} & \textbf{58.9} & \textbf{74} & \textsc{GPT-4o} + \fancynameshort(ours) & 8 & \textbf{56.9} & 60.7 & 68.2 \\
        \arrayrulecolor{gray!40}\hline \arrayrulecolor{black}

        \textsc{InternVL 2.5-78B}~\cite{chen2024internvl} & 8 & 55.7 & 57.3 & 74.0 & \textsc{InternVL 2.5-78B} & 8 & 52.6 & 55.5 & 55.9 \\
        \rowcolor{gray!10}
        \textsc{InternVL 2.5-78B} + \fancynameshort(ours) & 8 & \textbf{58.0} & \textbf{61.5} & \textbf{74.0}
        & \textsc{InternVL 2.5-78B} + \fancynameshort(ours) & 8 & \textbf{57.7} & \textbf{57.5} & \textbf{59.0} \\

        \midrule
        \textsc{GPT-4o} & 32 & 53.8 & 56.5 & 74.0 & {\textsc{GPT-4o}} & 32 & 55.2 & 61.0 & 71.4 \\
        \rowcolor{gray!10}
        \textsc{GPT-4o} + \tstar & 32 & 55.3 & 58.8 & 72.0 & \textsc{GPT-4o} + \tstar & 32 & 55.2 & 61.6 & \textbf{72.6} \\
        \rowcolor{gray!10}
        \textsc{GPT-4o} + \fancynameshort(ours) & 32 & \textbf{54.2} & \textbf{60.0} & \textbf{76.0} & \textsc{GPT-4o} + \fancynameshort(ours)& 32 & \textbf{55.2} & \textbf{61.9} & 71.9 \\
        \arrayrulecolor{gray!40}\hline \arrayrulecolor{black}

        \midrule
        \color{gray} \textsc{LLaVA-OneVision-QWen2-78B-ov} & \color{gray} 32 & \color{gray} 59.3 & \color{gray} 63.9 & \color{gray} 77.4 & \color{gray} LLaVA-OneVision-78B & \color{gray} 32 & \color{gray} 60.0 & \color{gray} 62.2 & \color{gray} 66.3 \\
        \color{gray} \textsc{PLLaVA-34B} & \color{gray} 32 & \color{gray} 49.1 & \color{gray} 50.8 & \color{gray} 66.8 & \color{gray} \textsc{VideoLLaMA 2} & \color{gray} 32 & \color{gray} 57.6 & \color{gray} 59.9 & \color{gray} 62.4 \\
        \color{gray} \textsc{LLaVA-Video-78B-Qwen2} & \color{gray} 128 & \color{gray} 59.3 & \color{gray} 63.9 & \color{gray} 77.4 & \color{gray} \textsc{Oryx-1.5} & \color{gray} 128 & \color{gray} 59.3 & \color{gray} 65.3 & \color{gray} 67.3 \\
        \color{gray} \textsc{mPLUG-Owl3-7B} & \color{gray} 128 & \color{gray} 53.9 & \color{gray} 58.8 & \color{gray} 73.7 & \color{gray} \textsc{Aria-8x3.5B} & \color{gray} 256 & \color{gray} 58.8 & \color{gray} 67.0 & \color{gray} 67.6 \\
        \color{gray} \textsc{GPT-4o} (0513) & \color{gray} 256 & \color{gray} 61.6 & \color{gray} {66.7} & \color{gray} 76.8 & \color{gray} \textsc{Gemini-1.5-Pro (0615)} & \color{gray} 1/0.5 fps & \color{gray} 67.4 & \color{gray} 74.3 & \color{gray} 75.0 \\
        \bottomrule
    \end{tabular}
    \end{adjustbox}
    \caption{
\textbf{Downstream task evaluation results on two benchmarks.}
All accuracy scores (\%) in black are from our replication. We also cite the reported accuracy of SOTA models in \textcolor{gray}{gray} (noting that their settings may differ and results may not be reproducible), along with their number of frames used for QA inference, for full transparency.
}
\label{tab:qa_table}
\vspace{-1.25em}
\end{table*}
To demonstrate the advantages of \fancynameshort, we evaluate downstream video QA performance on \lvb and \videomme. As shown in Table~\ref{tab:qa_table}, videos are grouped by length into \textbf{Short}, \textbf{Medium}, and \textbf{Long} (15–3600s, up to 60 mins). \fancynameshort consistently achieves the highest accuracy in the long-video category across different frame counts and QA models. Compared to the baseline \tstar, incorporating our visual semantic logic relations (Figure~\ref{fig:first}) yields substantial gains. These results confirm that modeling visual-logical relations is key to effective QA on long videos.

%% file: sec/Analysis.tex
\vspace{-1em}
\section{Analysis}
\vspace{-0.5 em}
\subsection{Coverage Analysis of Semantic-Logical Relations}
To ascertain the practical applicability and coverage of our defined semantic-logical relations (spatial, temporal, attribute, and causal), we conducted an analysis of their detection across all queries in the LongVideoBench and VideoMME datasets. Our findings reveal a crucial insight: for every question posed within these extensive VQA benchmarks, our query analysis module successfully identified and mapped the query to at least one of the four defined logical relation types. This empirical result supports the completeness of our proposed relation set for interpreting the semantic and logical intent inherent in these VQA tasks.
\vspace{-0.5em}
\subsection{Time Complexity}
\vspace{-0.5em}
\begin{wrapfigure}{r}{0.48\textwidth}
  \centering
  \vspace{-1em}
  \includegraphics[width=\linewidth]{"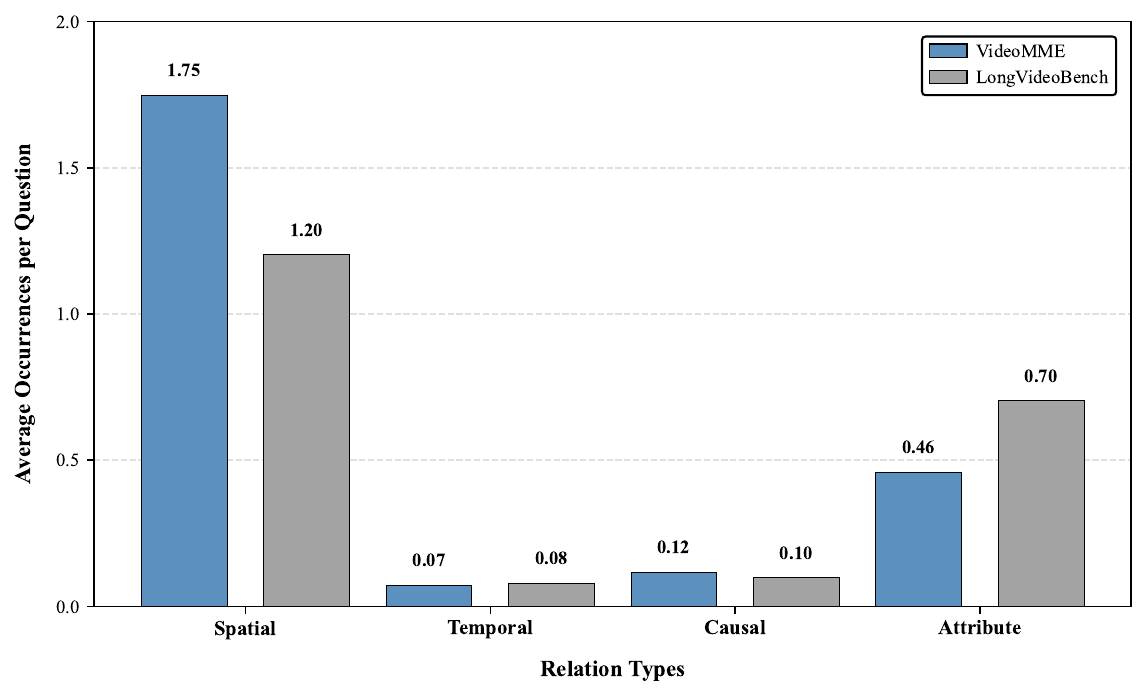"}
  \vspace{-1em}
  \caption{Average occurrences of detected semantic-logical relation types per question on the VideoMME and LongVideoBench datasets. Spatial relations are the most frequently identified, while all queries in both datasets triggered at least one of the four relation types.}
  \label{fig:logical_visual}
  \vspace{-1em}
\end{wrapfigure}

The proposed framework consists of two stages. First, VLMs such as \textsc{LlaVA-7B} and \textsc{GPT-4o} extract a semantic set $\mathcal{S}$ from a video $V$ with $n$ frames. $\mathcal{S}$ includes target objects, cue objects, and their relations, with their size constrained by prompt design. In the second stage, keyframe identification is performed via a heuristic search: $k$ candidates are iteratively selected using a scoring function $h(\cdot, \mathcal{S})$. The score distribution $\text{scores}[n]$ is dynamically refined using outputs from the \textsc{YOLO-World} detector.

Our analysis focuses on \textsc{YOLO-World} detections, the main computational bottleneck due to their reliance on deep neural networks. Reducing the number of detections improves efficiency without sacrificing accuracy. At each iteration, the detector processes $k$ selected frames to match objects and relations in $\mathcal{S}$, yielding $k$ detections. The search stops when all targets are found or the iteration budget $\min(1000,\ 0.1 \times V_t)$ (with $V_t$ as the video duration in seconds) is exhausted. In the worst case (e.g., videos with $>$10,000 frames and no matches), the cap is 1,000 iterations. Ideally, the evaluation function $h(\cdot,\mathcal{S})$ assigns high confidence to target frames, making the algorithm resemble top-$k$ selection over $n$ candidates in $\mathcal{O}(|\mathcal{S}|\log n)$ iterations~\cite{ye2025rethinking}, resulting in an average of $\mathcal{O}(|\mathcal{S}|k \log n)$ \textsc{YOLO-World} inferences.

Experimental results also demonstrate that integrating relational information into the search algorithm incurs negligible computational overhead compared to the baseline \tstar approach. On the \lvh benchmark, the average iteration count increases from 42.94 (\tstar) to 48.82 iterations, representing a modest 13.69\% rise in the time cost. 

\vspace{-0.5em}
\subsection{Ablation Study of Four Relations}
\vspace{-0.15em}

\begin{wraptable}{r}{0.54\textwidth}
\vspace{-2em}
    \centering
    \small
    \setlength{\tabcolsep}{3pt}
    \setlength\tabcolsep{3pt}
    \setlength\extrarowheight{1pt}
    \vspace{-1em}
    \begin{tabular}{c|lccc}
\hline
\multicolumn{1}{c|}{} & \multicolumn{1}{l|}{} & \multicolumn{3}{c}{\textbf{\lvb}} \\
\multicolumn{1}{c|}{\multirow{-2}{*}{\textbf{Logic Type}}}
 & \multicolumn{1}{c|}{\multirow{-2}{*}{\textbf{Method}}}
 & Precision $\uparrow$
 & Recall $\uparrow$
 & TC $\uparrow$
\\ \hline

% 1) Spatial
 & \tstar & 72.9 & 88.7 & 37.5 \\
\multirow{-2}{*}{\textit{Spatial}} 
 & \cellcolor{gray!20}\textbf{\fancynameshort(ours)} 
 & \cellcolor{gray!20}\textbf{73.6} 
 & \cellcolor{gray!20}\textbf{91.4} 
 & \cellcolor{gray!20}\textbf{45.5} 
\\ \hline

% 2) Attribute
 & \tstar & 71.8 & 87.6 & 38.5 \\
\multirow{-2}{*}{\textit{Attribute}} 
 & \cellcolor{gray!20}\textbf{\fancynameshort(ours)} 
 & \cellcolor{gray!20}\textbf{72.7} 
 & \cellcolor{gray!20}\textbf{90.9} 
 & \cellcolor{gray!20}\textbf{42.1} 
\\ \hline

% 3) Temporal (Time)
 & \tstar & 76.7 & 89.2 & 37.3 \\
\multirow{-2}{*}{\textit{Time}} 
 & \cellcolor{gray!20}\textbf{\fancynameshort(ours)} 
 & \cellcolor{gray!20}\textbf{77.5} 
 & \cellcolor{gray!20}\textbf{92.5} 
 & \cellcolor{gray!20}36.1
\\ \hline

% 4) Causal
 & \tstar & 74.7 & 92.4 & 38.6 \\
\multirow{-2}{*}{\textit{Casual}} 
 & \cellcolor{gray!20}\textbf{\fancynameshort(ours)} 
 & \cellcolor{gray!20}74.7 
 & \cellcolor{gray!20}\textbf{93.8} 
 & \cellcolor{gray!20}\textbf{39.6} 
\\ \hline

\end{tabular}
    \caption{Comparison of our method (\textbf{VSLS}) with the baseline across four logic relation types on \lvb. 
    \textbf{Precision}: SSIM Precision; 
    \textbf{Recall}: SSIM Recall; 
    \textbf{TC}: Temporal Coverage.
    }
    \label{tab:searching_utility}
    \vspace{-1em}
\end{wraptable}

Figure~\ref{fig:logical_visual} illustrates the distribution of four logic relation types across \lvb and \videomme datasets, where \textit{spatial} relations predominate, followed by \textit{attribute} relations. In Table~\ref{tab:searching_utility}, we extract samples containing different relation types from \lvb to compare the object detection-based \tstar method with our \fancynameshort approach. Experimental results demonstrate that \fancynameshort achieves significant improvements across both image similarity metrics (SSIM Precision and SSIM Recall). Additionally, temporal coverage shows marked enhancement for \textit{attribute}, \textit{spatial}, and \textit{causal} relations, with \textit{spatial} relations exhibiting the most substantial improvement (21.3\% increase over \tstar). For the \textit{time} relation category, we observe a slight decrease in temporal coverage, which may be attributed to the relative scarcity of time relation samples in the dataset, limiting the opportunity to demonstrate the advantages of \fancynameshort. Nevertheless, Figure~\ref{fig:first} provides visual evidence of how effectively leveraging time relations can facilitate downstream question-answering tasks.

%% file: sec/relatedwork.tex
\vspace{-1em}
\section{Related Work}
\vspace{-0.5em}

\noindent \textbf{Challenges in Long Video Understanding:} Long video understanding is inherently more challenging than short-video or image-based tasks due to its rich temporal dynamics and massive redundancy~\cite{qian2024streaming, zeng2024timesuite, yu2019activityqa}. The large number of frames increases both memory and computational requirements, making straightforward dense sampling infeasible. Moreover, crucial events may span distant timestamps, demanding high-capacity models to capture long-range dependencies~\cite{ranasinghe2025understanding, shi2024unlocking, chen2024rextime,weng2024longvlm}. Meanwhile, the diverse and continuous visual content raises noise and distractors; thus, strategies to effectively locate or distill essential parts of the video are of primary importance~\cite{84, cheng2024videollama2advancingspatialtemporal, xu2023retrieval, ye2025rethinking}.

\noindent \textbf{Existing Solutions} based on VLMs typically share three core ideas: 1) \emph{video sampling or retrieval} for efficiency, 2) \emph{multi-stage or interactive reasoning} to handle complex questions, and 3) \emph{compact representation} to accommodate the VLM's limited context window. For instance, retrieval-based pipelines partition a video into segments and employ a learned or rule-based retriever to identify the relevant chunks before passing them to a VLM~\cite{pan2023retrieving, choudhury2023zero, rohan2025video}. Other lines of research compress each frame into minimal tokens to reduce computational overhead~\cite{li2024llamavid, chen2024llavolta, song2024less}, or adopt a streaming mechanism to propagate memory representations along the temporal axis~\cite{qian2024streaming, wu2022memvit,liu2024llava}. Beyond these efficiency-oriented approaches, LLM/VLM-as-planner frameworks factorize the process into a series of perception queries, enabling an agent to fetch additional frame-level details if needed~\cite{wang2024videoagent, zhang2024omagent, liao2024videoinsta}.

%% file: sec/conclusion.tex
\vspace{-1em}
\section{Conclusion}
\vspace{-0.5em}
In this paper, we present \texttt{Visual} \texttt{Semantic}-\texttt{Logical} \texttt{Search} (\fancynameshort), a novel framework that efficiently selects semantically keyframes for long video understanding by decomposing logical relationships between textual queries and visual elements. \fancynameshort based on four defined logical dependencies (spatial co-occurrence, temporal proximity, attribute dependency, and causal order), significantly outperforms existing methods while sampling only 1.4\% of video frames. The 8.7\% improvement in \textsc{GPT-4o}'s long video QA accuracy demonstrates that query-guided visual semantic logic search effectively bridges the gap between textual queries and visual content. \fancynameshort's plug-and-play nature enables seamless integration with existing pipelines, making it practical for real-world applications. Future work could consider more logical relations, learnable search methods, enhancing interpretability, and exploring more downstream tasks.

%% file: sec/appendix.tex
\section{Theoretical Underpinnings of Relation Categories}
\label{Theoretical}
Our choice of the four relation categories—--\textit{spatial, temporal, attribute, and causal}—--is grounded in foundational concepts from linguistics and logic. While achieving absolute ``completeness'' in describing the infinite complexity of the real world is a formidable challenge, this selection aims to describe core aspects of events, states, and the way humans conceptualize and communicate them.

\subsection{Linguistic Grounding}
\begin{description}[style=unboxed,leftmargin=0pt]
    \item[Semantic Roles and Case Grammar:] Theories like Fillmore's Case Grammar \cite{fillmore1968case} analyze sentences in terms of semantic roles that nominals play in relation to the verb (the event).
        \begin{itemize}[label=\textbullet, leftmargin=*]
            \item \textbf{Spatial relations} directly correspond to roles like \textit{Locative} (the location of an event or state) or \textit{Path} (the trajectory of motion).
            \item \textbf{Temporal relations} align with \textit{Temporal} roles, specifying when an event occurs or its duration.
            \item \textbf{Attributes} describe the properties of entities (participants) involved in these roles. While not direct case roles for verbs, they are fundamental for identifying and characterizing the ``who'' and ``what'' (e.g., Agent, Patient, Theme, Instrument) that possess these attributes during an event.
            \item \textbf{Causal relations} are central to understanding agency and event structure. Roles like \textit{Agent} (the instigator of an action) or \textit{Cause} (the non-volitional trigger of an event) highlight the importance of causality in linguistic descriptions of events.
        \end{itemize}

    \item[Lexical Semantics and Event Structure:] Works in lexical semantics (e.g., following Pustejovsky \cite{cohen1968universals} on the generative lexicon, or Talmy \cite{talmy2000toward} on cognitive semantics) often decompose event meaning into fundamental components. Talmy \cite{talmy2000toward}, for instance, extensively discusses how language structures concepts like space, time, and force dynamics (which inherently relate to causality). Events are situated in space and time, involve entities with specific attributes, and are often linked through causal chains (e.g., one action causing another, or an agent causing a change of state).

    \item[Discourse Relations:] Theories like Rhetorical Structure Theory (RST) \cite{mann1988rhetorical} identify relations that bind textual units together. Many of these fundamental relations are inherently temporal (e.g., \textit{Sequence}), causal (e.g., \textit{Cause, Result, Purpose}), or involve describing entities and their settings (which encompasses spatial and attributive information, often under relations like \textit{Elaboration} or \textit{Background}). This suggests that these four categories capture essential elements for constructing coherent descriptions and explanations, a core function of Video Question Answering (VQA).
\end{description}

\subsection{Logical Grounding}
\begin{description}[style=unboxed,leftmargin=0pt]
    \item[Predicate Logic and Knowledge Representation:] In formal logic and AI knowledge representation (e.g., Sowa \cite{sowa2000knowledge}), events and states are often represented using predicates with arguments that specify participants, locations, times, and properties. A typical event representation might implicitly or explicitly include \texttt{Location(event, place)}, \texttt{Time(event, time\_interval)}, \texttt{HasProperty(entity, attribute\_value)}, and relations like \texttt{Causes(event1, event2)}. Our four categories provide a high-level abstraction over these common predicate types.

    \item[Modal and Specialized Logics:]
        \begin{itemize}[label=\textbullet, leftmargin=*]
            \item \textbf{Temporal Logic} is specifically designed to reason about propositions qualified in terms of time.
            \item \textbf{Spatial Logic} deals with reasoning about spatial properties and relations between entities.
            \item Logics of \textbf{Action and Causality} (e.g., situation calculus, event calculus, or Pearl's work on causality \cite{neuberg2003causality}) explicitly model how actions bring about changes and the causal dependencies between events.
        \end{itemize}
\end{description}

\subsection{Pragmatic Completeness for VQA}
From a pragmatic standpoint, particularly for VQA, these four relations address the core ``Wh-questions'' humans often ask to understand a scene or event:
\begin{itemize}[label=\textbullet, leftmargin=*]
    \item \textbf{What/Who?} (Identifies objects/entities, often distinguished by their \textbf{attributes})
    \item \textbf{Where?} (Answered by \textbf{spatial} relations)
    \item \textbf{When?} (Answered by \textbf{temporal} relations)
    \item \textbf{Why/How did it happen?} (Often answered by \textbf{causal} relations or a sequence of events linked temporally and spatially)
\end{itemize}
While more fine-grained relations (as in Action Genome) undoubtedly provide deeper semantic detail, our chosen set aims to provide a foundational, yet computationally manageable, framework for keyframe selection based on the most common semantic and logical inferences required for a broad range of video queries. They represent a level of abstraction that is both meaningful for human queries and feasible for current visual-language models to parse and verify.

In essence, these categories are not arbitrary but reflect fundamental dimensions along which events and states are structured, perceived, and communicated in language and reasoned about in logic. We believe they offer a robust and broadly applicable framework for the task at hand.
\section{Performance}
Long-form video understanding presents unique challenges due to the complexity of temporal dynamics and cross-modal interactions in extended durations (900-3,600 seconds). Our comprehensive evaluation of the LVB-XL benchmark reveals significant performance gaps between existing approaches. While large-scale models like \textsc{GPT-4o} (32 frames) and \textsc{InternVL 2.5-78B} (16 frames) have demonstrated competence in short-video tasks, their direct application to long-form content (marked by circle sizes proportional to model parameters) yields suboptimal results (53.8\% and 56.5\% accuracy respectively). 

Our \texttt{Visual} \texttt{Semantic}-\texttt{Logical} \texttt{Search} (\fancynameshort) framework addresses these limitations. This advancement enables consistent performance improvements across different architecture scales, elevating \textsc{GPT-4o} to 54.2\% (+0.4pp) and achieving a remarkable 62.4\% (+5.9pp) for \textsc{InternVL 2.5-78B} on this benchmark. The comparative analysis further suggests that \fancynameshort's gains become particularly pronounced when processing longer visual sequences, highlighting its effectiveness in modeling extended temporal contexts.

\section{Analysis of the Impact of Search Frame Count}
\begin{figure}[ht] 
\centering 
\includegraphics[width=0.75\linewidth]{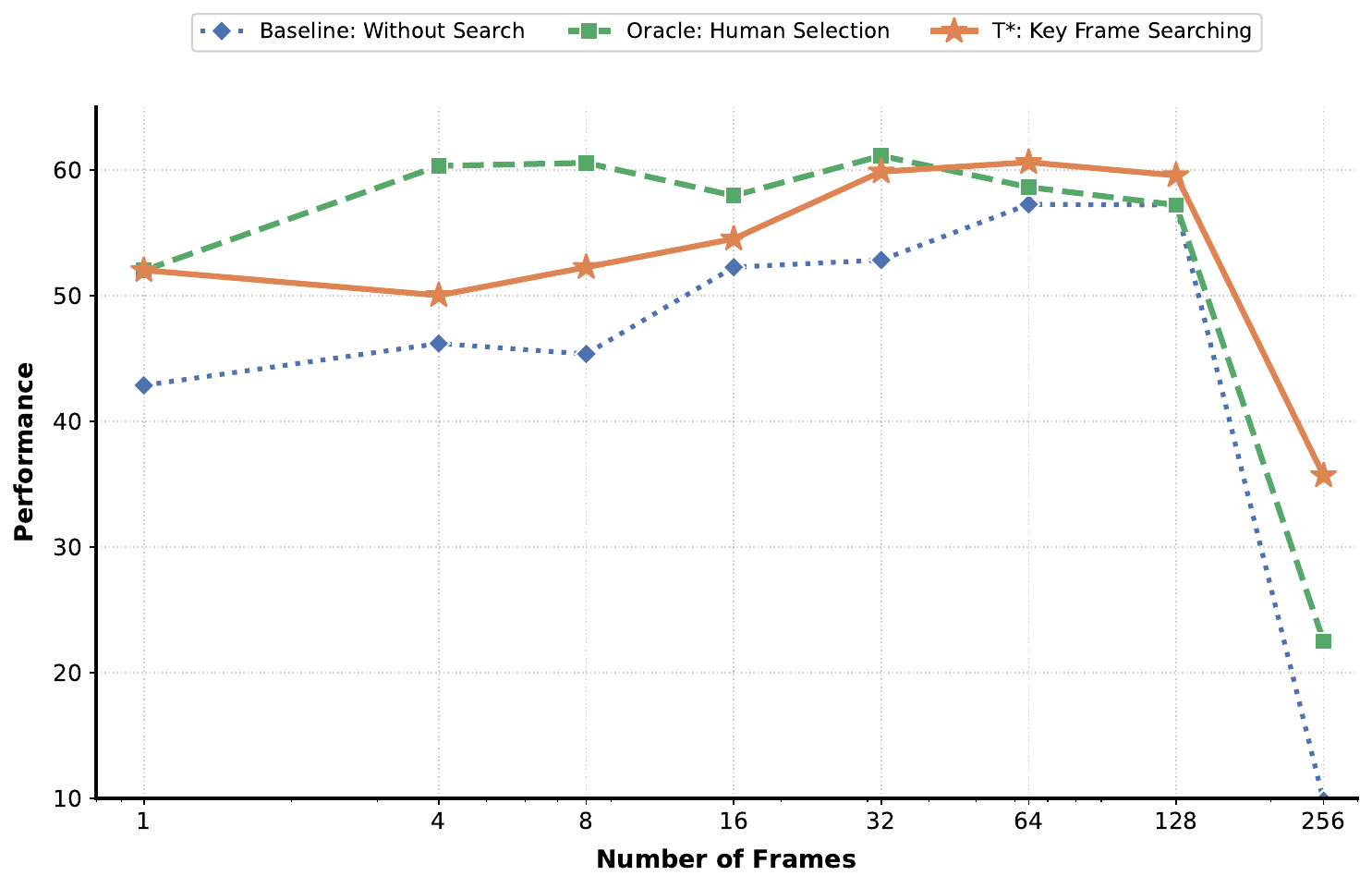} 
\caption{Performance improvement with increasing search frames. \fancynameshort consistently enhances accuracy and reaches near-human oracle performance at 64 frames.}
\label{fig:frames_vs_acc} 
\end{figure}

This section investigates the impact of the number of search frames on the performance of our Visual Language Models (VLMs) in the context of \lvb.

Figure~\ref{fig:frames_vs_acc} in the \tstar framework study empirically demonstrates the non-monotonic relationship between input frame quantity and model accuracy on the \lvb XL benchmark. Through systematic experimentation across 18 state-of-the-art VLMs, this visualization reveals a critical phenomenon: excessive frame inputs degrade performance for models lacking temporal redundancy mitigation mechanisms.

\section{Details of Datasets}
\label{app:details}

\subsection{Details of \videomme}
The \videomme (Video Multi-Modal Evaluation) dataset represents the first comprehensive benchmark tailored to assess the capabilities of Vision-Language Models (VLMs) in video understanding. Aiming to address limitations in existing benchmarks, it emphasizes diversity, temporal complexity, and multi-modal integration while ensuring high-quality human annotations. The dataset contains 900 carefully curated videos across six primary domains—Knowledge, Film and Television, Sports Competition, Artistic Performance, Life Record, and Multilingual—with 30 fine-grained subcategories such as astronomy, esports, and documentaries. These videos vary significantly in duration, ranging from short clips (11 seconds) to long-form content (up to 1 hour), enabling robust evaluation across temporal scales.

Each video is paired with expert-annotated multiple-choice questions (2,700 QA pairs in total), rigorously validated to ensure clarity and reliance on visual or multi-modal context. Questions span 12 task types, including action recognition, temporal reasoning, and domain-specific knowledge, with a focus on scenarios where answers cannot be inferred from text alone. To quantify temporal complexity, the dataset introduces certificate length analysis, revealing that answering questions often requires understanding extended video segments (e.g., median lengths of 26 seconds for short videos and 890.7 seconds for long videos), surpassing the demands of prior benchmarks like \textsc{EgoSchema}.

\videomme serves as a universal benchmark, applicable to both image- and video-focused MLLMs, and exposes key challenges for future research. These include improving architectures for long-sequence processing, developing datasets for complex temporal reasoning, and enhancing cross-modal alignment. By providing a rigorous evaluation framework, \videomme aims to drive progress toward MLLMs capable of understanding dynamic, real-world scenarios.

\subsection{Details of \lvb}
The \lvb benchmark pioneers the evaluation of long-context interleaved video-language understanding in VLMs, addressing critical gaps in existing benchmarks through its focus on detailed retrieval and temporal reasoning over hour-long multimodal inputs. Designed to overcome the "single-frame bias" prevalent in prior video benchmarks, the novel referring reasoning paradigm enables models to locate and analyze specific contexts within extended sequences. The data set comprises 3,763 web-sourced videos that span various themes - movies, news, life vlogs, and knowledge domains (including art, history, and STEM) - with durations progressively grouped into four levels: 8-15 seconds, 15-60 seconds, 3-10 minutes, and 15-60 minutes. Each video is paired with aligned subtitles, forming interleaved multimodal inputs that mimic real-world viewing scenarios.

The benchmark features 6,678 human-annotated multiple-choice questions categorized into 17 fine-grained task types across two levels: Perception (requiring object/attribute recognition in single scenes) and Relation (demanding temporal/causal reasoning across multiple scenes). Questions incorporate explicit referring queries (e.g., "When the woman descends the rocky hill...") that anchor reasoning to specific video moments, with an average question length of 43.5 words to ensure precision. Temporal complexity is quantified through duration-grouped analysis, where models must process up to 256 frames (at 1 fps) for hour-long videos, significantly exceeding the demands of predecessors like \textsc{EgoSchema} (180s videos).

\subsection{Details of \lvh}
The \lvh benchmark establishes the first comprehensive evaluation framework for temporal search in long-form video understanding, addressing critical limitations in existing synthetic needle-in-haystack benchmarks through real-world video annotations and multi-dimensional evaluation metrics. Designed to assess models' ability to locate minimal keyframe sets (typically 1-5 frames) from hour-long videos containing tens of thousands of frames, the dataset comprises 3,874 human-annotated instances spanning 150 hours of video content across two distinct categories: egocentric videos from \textsc{Ego4D} (101 hours) and allocentric videos from \lvb (57.7 hours).

Organized into \textsc{Haystack-Ego4D} and \hlvb subsets, the benchmark features videos averaging 24.8 minutes in length (max 60 minutes) with 44,717 frames per video. Each instance contains:
\begin{itemize}
    \item Expert-curated multi-choice questions requiring temporal reasoning (15.9 questions/video);
    \item Human-annotated keyframe sets (4.7 frames/question for egocentric, 1.8 frames/question for allocentric);
    \item Temporal and visual similarity metrics for precise search evaluation.
\end{itemize}

\subsection{Details of \textsc{Ego-4D}}

The \textsc{Ego4D} (Egocentric Computer Vision Benchmark) dataset establishes a transformative foundation for advancing research in first-person visual perception through unprecedented scale, diversity, and multi-modal integration. Designed to overcome limitations in existing egocentric datasets, it captures 3,670 hours of unscripted daily activities from 931 participants across 74 global locations and 9 countries, spanning household, workplace, leisure, and outdoor scenarios. The dataset features 30+ fine-grained activity categories including carpentry, social gaming, and meal preparation, with videos ranging from brief interactions (8-minute clips) to extended continuous recordings (up to 10 hours), enabling comprehensive analysis of long-term behavioral patterns.

Each video is enriched with multi-modal annotations totaling 3.85 million dense textual narrations (13.2 sentences/minute), coupled with 3D environment meshes, eye gaze tracking, stereo vision, and synchronized multi-camera views. Rigorous privacy protocols ensure ethical data collection, with 612 hours containing unblurred faces/audio for social interaction studies. The benchmark suite introduces five core tasks organized across temporal dimensions:

\begin{itemize}
    \item \textbf{Episodic Memory}: Temporal localization of natural language queries (74K instances) and 3D object tracking using Matterport scans;
    \item \textbf{Hand-Object Interaction}: State change detection (1.3M annotations) with PNR (point-of-no-return) temporal localization;
    \item \textbf{Social Understanding}: Audio-visual diarisation (2,535h audio) and gaze-directed communication analysis;
    \item \textbf{Action Forecasting}: Anticipation of locomotion trajectories and object interactions.
\end{itemize}

Quantitative analysis reveals the dataset's complexity: hand-object interactions involve 1,772 unique verbs and 4,336 nouns, while social scenarios contain 6.8 participant interactions per minute on average. Multi-modal fusion experiments demonstrate performance gains, with 3D environment context improving object localization accuracy by 18.7\% compared to RGB-only baselines. State-of-the-art models achieve 68.9\% accuracy in action anticipation tasks, yet struggle with long-term forecasting (41.2\% accuracy for 5s predictions), highlighting critical challenges in temporal reasoning.

\textsc{Ego4D}'s unique integration of egocentric video with complementary modalities (IMU data in 836h, gaze tracking in 45h) enables novel research directions in embodied AI and augmented reality. The dataset exposes fundamental limitations in current architectures, particularly in processing hour-long video contexts and synthesizing cross-modal signals—only 23\% of tested models effectively utilized audio-visual synchronization cues. By providing standardized evaluation protocols and curated challenge subsets, \textsc{Ego4D} serves as a universal testbed for developing perceptive systems capable of understanding persistent 3D environments and complex human behaviors.

\section{Detailed Algorithm}
\label{app:detailed_algorithm}
The detailed \fancynameshort algorithm is represented in Algorithm~\ref{alg:slt_search}.

\begin{algorithm*}
\caption{The completed Visual Semantic-Logical Search}\label{alg:slt_search}
\footnotesize
\SetAlgoNlRelativeSize{0}
\SetNlSty{}{}{:}
\SetKwFunction{SLTSearch}{SemanticLogicalTemporalSearch}
\SetKwFunction{DetectObjects}{DetectObjects}
\SetKwFunction{CalcBaseScore}{CalculateBaseScore}
\SetKwFunction{UpdateScores}{UpdateScores}
\SetKwFunction{CheckSpatial}{CheckSpatialRelationship}
\SetKwFunction{CheckTemporal}{CheckTemporalRelationship}
\SetKwFunction{CheckCausal}{CheckCausalRelationship}
\SetKwFunction{CheckAttribute}{CheckAttributeRelationship}
\SetKwFunction{DiffuseScores}{DiffuseScores}
\SetKwFunction{Normalize}{NormalizeDistribution}
\SetKwInOut{Input}{Input}\SetKwInOut{Output}{Output}
\SetKwProg{Fn}{Function}{:}{end} 

\Fn{\SLTSearch{$V, Q, K, \Delta_t, \tau, \alpha, \gamma$}}{
    $\mathcal{O}, \mathcal{R} \gets \text{ParseQuestion}(Q)$ \tcp*[r]{\textcolor{commentgray}{Extract key/cue objects and relationships}}
    $P \gets \text{Uniform}, B \gets |V|, S \gets \emptyset, N_v \gets |V|$ \tcp*[r]{\textcolor{commentgray}{Initialize distribution and state}}
    
    \While{$B > 0$ \textbf{and} $|\mathcal{O}| > 0$}{
        $k \gets \lfloor\sqrt{B}\rfloor$, $G \gets \Grid(\Sample(P, k^2))$ \tcp*[r]{\textcolor{commentgray}{Adaptive grid sampling}}
        $\Omega \gets \DetectObjects(G)$ \tcp*[r]{\textcolor{commentgray}{Detect objects in sampled frames}}
        
        \ForEach{$g \in G$}{
            $C_g \gets \CalcBaseScore(\Omega[g])$ \tcp*[r]{\textcolor{commentgray}{Base detection confidence}}
            \ForEach{$r \in \mathcal{R}$}{
                \uIf{$r.\text{type} = \text{Spatial}$}{
                    $C_g \gets C_g + \alpha\gamma_{\text{spatial}} \cdot \CheckSpatial(r, \Omega[g])$
                }
                \uElseIf{$r.\text{type} = \text{Temporal}$}{
                    $C_g \gets C_g + \alpha\gamma_{\text{time}} \cdot \CheckTemporal(r, \Omega, \Delta_t)$
                }
                \uElseIf{$r.\text{type} = \text{Causal}$}{
                    $C_g \gets C_g + \alpha\gamma_{\text{causal}} \cdot \CheckCausal(r, \Omega)$
                }
                \uElseIf{$r.\text{type} = \text{Attribute}$}{
                    $C_g \gets C_g + \alpha\gamma_{\text{attr}} \cdot \CheckAttribute(r, \Omega[g], \tau)$
                }
            }
            $\UpdateScores(S, g, C_g)$ \tcp*[r]{\textcolor{commentgray}{Update global score registry}}
        }
        
        $\DiffuseScores(S, w)$ \tcp*[r]{\textcolor{commentgray}{Temporal context propagation}}
        $P \gets \Normalize(S)$, $B \gets B - k^2$ \tcp*[r]{\textcolor{commentgray}{Update sampling distribution}}
        
        \ForEach{$g \in \TopK(S, K)$}{
            \If{$\Omega[g] \cap \mathcal{O} \neq \emptyset$}{
                $\mathcal{O} \gets \mathcal{O} \setminus \Omega[g]$\tcp*[r]{\textcolor{commentgray}{Remove identified key objects}}
            }
        }
    }
    \Return $\TopK(S, K)$ \tcp*[r]{\textcolor{commentgray}{Return top-K keyframes}}
}
\end{algorithm*}

\subsection{Algorithm Overview and Core Components}

The algorithm operates as an adaptive search framework that intelligently explores video content (represented as set $V$) to locate frames matching semantic-logical query requirements ($Q$). Unlike traditional linear search methods, it employs a probabilistic sampling strategy that dynamically adjusts based on confidence scores from multiple relationship types.

\paragraph{Initialization Phase}

The process begins by parsing the input query $Q$ into two fundamental components:
\begin{itemize}
    \item $\mathcal{O}$: A set of key objects or entities to identify
    \item $\mathcal{R}$: A collection of relationships (spatial, temporal, causal, and attribute) that must be satisfied
\end{itemize}

The algorithm initializes with a uniform probability distribution ($P$) across all video frames, establishing a budget ($B$) equivalent to the total number of frames ($|V|$), and creating an empty score registry ($S$) to track confidence values. This approach ensures unbiased initial exploration before evidence-guided refinement.

\paragraph{Adaptive Sampling Strategy}

Rather than exhaustively processing every frame, the algorithm employs a square-root scaling sampling strategy where $k = \lfloor\sqrt{B}\rfloor$ determines the sampling density. This provides a mathematical balance between exploration breadth and computational efficiency. The Grid function organizes sampled frames into a structured representation that preserves spatial-temporal relationships, facilitating subsequent relationship analysis.

\paragraph{Multi-modal Object Detection}

The DetectObjects function applies state-of-the-art computer vision techniques to identify objects within each sampled frame. This step leverages deep neural networks pre-trained on diverse visual datasets, enabling recognition of a wide range of entities with their corresponding confidence scores and spatial locations within frames.

\paragraph{Score Propagation and Distribution Update}

The $\mathrm{DiffuseScores}$ function implements a temporal context propagation mechanism that spreads confidence values to neighboring frames, acknowledging that relevant content likely extends beyond individual frames. This diffusion creates a smoothed confidence landscape that guides subsequent sampling.

After each iteration, the algorithm normalizes the accumulated scores to form an updated probability distribution, focusing future sampling on promising regions while maintaining exploration potential in unexamined areas.

\paragraph{Convergence Criteria and Termination}

The search continues until either:
\begin{itemize}
    \item The sampling budget ($B$) is exhausted, indicating comprehensive coverage of the video content
    \item All target objects ($\mathcal{O}$) have been successfully identified at satisfactory confidence levels
\end{itemize}

This dual-termination approach balances thoroughness with efficiency, preventing unnecessary computation once objectives are met.

\paragraph{Result Generation}

The algorithm concludes by returning the top-K frames with the highest confidence scores, representing the most relevant video segments that satisfy the semantic-logical query requirements. These keyframes provide a concise summary of the content matching the user's information needs.

\subsection{Implementation Considerations}

The algorithm's performance depends on several configurable parameters:
\begin{itemize}
    \item $\Delta_t$: Temporal window size for relationship analysis
    \item $\tau$: Confidence threshold for attribute matching
    \item $\alpha$: Global relationship influence factor
    \item $\gamma$: Type-specific relationship weights
\end{itemize}

These parameters can be tuned based on application requirements, video characteristics, and computational constraints. The algorithm's modular design allows for straightforward substitution of specific component implementations (e.g., different object detectors or relationship checkers) without altering the overall framework.

\subsection{Computational Complexity Analysis}

The time complexity scales with $O(\sqrt{N})$ where $N$ is the total number of frames, significantly improving upon linear approaches. Space complexity remains $O(N)$ to maintain the probability distribution and score registry. The algorithm intelligently balances exploration and exploitation through its adaptive sampling approach, making it particularly suitable for large-scale video analysis tasks where exhaustive processing would be prohibitive.

\subsection{Technical Implementation Details}

\paragraph{Object Detection and Feature Extraction}

To achieve real-time performance, the object detection module utilizes pre-trained deep convolutional neural network architectures, particularly variants based on \textsc{Fast R-CNN} and \textsc{YOLO} series. The system employs a two-stage detection strategy:

\begin{itemize}
\item \textbf{Preliminary Detection}: Using lightweight models to rapidly identify potential regions;
\item \textbf{Fine-grained Classification}: Applying more sophisticated models for detailed classification on high-confidence regions.
\end{itemize}

The feature extraction process leverages self-attention mechanisms from Visual Transformers (ViT), generating rich semantic embeddings robust to various visual variations such as scale, rotation, and illumination. Each identified object is associated with a feature vector $f_i \in \mathbb{R}^d$, where $d=512$ represents the dimensionality of the embedding space.

\paragraph{Mathematical Formulations for Relationship Assessment}

The evaluation of various relationship types is based on precise mathematical definitions:

\paragraph{Spatial Relationships} Given bounding boxes $B_i=(x_i, y_i, w_i, h_i)$ and $B_j=(x_j, y_j, w_j, h_j)$ for two objects, the confidence for a spatial relationship $r_{spatial}$ is calculated as:

\begin{equation}
C_\text{spatial}(B_i, B_j, r) = \phi_{r}(B_i, B_j) \cdot \psi(B_i) \cdot \psi(B_j),
\end{equation}

where $\phi_{r}$ is a relationship-specific compatibility function and $\psi$ is the object detection confidence. For example, the compatibility for a "contains" relationship is defined as:

\begin{equation}
\phi_\text{contains}(B_i, B_j) = \frac{\text{IoU}(B_i, B_j)}{\text{Area}(B_j)}.
\end{equation}

\paragraph{Temporal Relationships} Temporal relationships are calculated by evaluating object behavior patterns across a sequence of frames $\{F_t, F_{t+1}, ..., F_{t+\Delta_t}\}$:

\begin{equation}
C_\text{temporal}(O_i, O_j, r, \Delta_t) = \prod_{k=0}^{\Delta_t-1} T_r(O_i^{t+k}, O_j^{t+k+1}),
\end{equation}

where $T_r$ is a relationship-specific temporal transition matrix and $O_i^t$ represents the state of object $i$ at time $t$.

\paragraph{Causal Relationships} Causal relationships utilize a Bayesian network framework to compute conditional probabilities:

\begin{equation}
C_\text{causal}(E_i, E_j) = P(E_j | E_i) \cdot \log\frac{P(E_j | E_i)}{P(E_j)},
\end{equation}

where $E_i$ and $E_j$ represent the presumed cause event and effect event, respectively.

\paragraph{Attribute Relationships} Attribute evaluation employs cosine similarity metrics between feature vectors and attribute prototypes:

\begin{equation}
C_\text{attr}(O_i, a) = \max(0, \cos(f_i, p_a) - \tau),
\end{equation}

where $p_a$ is the prototype vector for attribute $a$ and $\tau$ is the minimum similarity threshold.

\paragraph{Score Propagation Algorithm}

Temporal score propagation is implemented through a weighted diffusion process, analogous to heat diffusion on a graph structure:

\begin{equation}
S'(t) = S(t) + \sum_{k \in \mathcal{N}(t)} w_{k,t} \cdot S(k),
\end{equation}

where $\mathcal{N}(t)$ represents the temporal neighborhood of frame $t$, and $w_{k,t}$ is a weight based on temporal distance, defined as:

\begin{equation}
w_{k,t} = \exp\left(-\frac{|k-t|^2}{2\sigma^2}\right),
\end{equation}

where $\sigma$ controls the diffusion range.

\paragraph{Adaptive Sampling Optimization}

The sampling strategy is further improved through a dynamically adjusted Thompson sampling method, modeling the probability distribution $P$ as a Beta distribution with shape parameters updated through previous observations:

\begin{equation}
P(t) \sim \text{Beta}(\alpha_t + \sum_{i} S_i(t), \beta_t + n - \sum_{i} S_i(t)),
\end{equation}

where $\alpha_t$ and $\beta_t$ are prior hyperparameters and $n$ is the total number of observations.

\subsection{Practical Application Examples}

In practical visual search scenarios, the algorithm processes complex queries such as "\textit{a person wearing a blue shirt sits down at a table and then picks up a coffee cup}":
\begin{itemize}
\item Query parsing identifies key objects (\texttt{person, shirt, table, coffee cup}) and relationships (\texttt{blue attribute, sitting action, temporal before-after relation, spatial proximity});
\item Adaptive sampling selects representative frames from the video;
\item Multi-relationship evaluation integrates various sources of evidence;
\item Score propagation establishes a unified confidence landscape across related frame sets;
\item Result generation provides a concise summary of the most relevant segments in the video.
\end{itemize}

This semantic-logical-temporal search framework represents a significant advancement in multimodal content retrieval, enabling natural language queries that incorporate complex relationships across objects, time, and causal chains.

\subsection{System Specifications for Reproductivity}
\label{app:system_specifications}
Our experiments were conducted on high-performance servers, each equipped with either an Intel(R) Xeon(R) Platinum 8378A CPU @ 3.00GHz or an Intel(R) Xeon(R) Platinum 8358P CPU @ 2.60GHz, 1TB of RAM, and 4/6 NVIDIA A800 GPUs with 80GB memory. Machines with 4 GPUs are configured with the SXM4 version, while those with 6 GPUs use the PCIe version. The software environment included \textit{Python} 3.11, \textit{PyTorch} 2.4, and \textit{NCCL} 2.21.5 for reproductivity.

\section{Case Study of \fancynameshort\ Keyframe Selection}
\label{sec:case_study}

\begin{figure}[htp]
\centering
\includegraphics[width=0.8\textwidth]{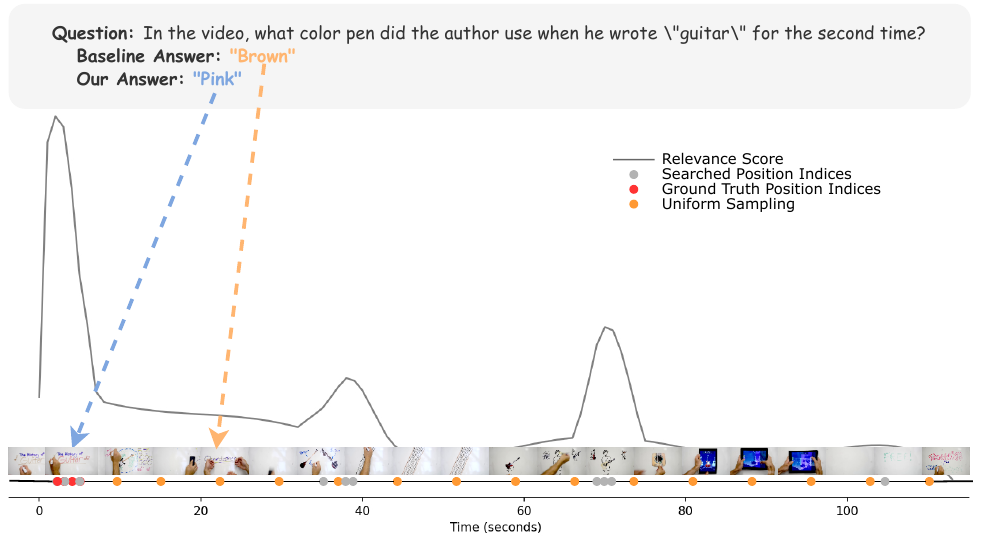}
\caption{Qualitative comparison of frame selection strategies demonstrates \fancynameshort's ability to pinpoint query-critical moments (e.g., the subject presenting pink objects) with temporal precision, while baseline approaches exhibit color misinterpretation (brown) due to suboptimal frame choices. \fancynameshort maintains superior temporal diversity and content relevance, effectively avoiding the redundant selections observed in comparative methods.}
\label{fig:case_study}
\end{figure}

As shown in Figure~\ref{fig:case_study}, the \fancynameshort framework demonstrates its effectiveness through a video question-answering case study involving temporal handwriting analysis. The experiment focuses on distinguishing between two sequential events: a brown pen writing "guitar" at 2 seconds and a pink pen rewriting the same word at 3 seconds, with the query requiring identification of the second occurrence's pen color.

\fancynameshort's analytical process unfolds through three interpretable phases:

\begin{itemize}
\item \textbf{Semantic Logic Extraction}: Identifies core visual entities (\textit{handwritten text}, \textit{pen}, \textit{paper}) and constructs temporal relationships through triplet formulation: (\textit{text}, \textit{time}, \textit{pen}), establishing the framework for tracking writing instrument changes;

\item \textbf{Temporal Relevance Scoring}: The gray relevance curve reveals precise temporal localization, with peak scores  aligning perfectly with ground truth positions at 2s and 3s, contrasting sharply with baseline methods' random fluctuations;

\item \textbf{Search Pattern Visualization}: Demonstrates \fancynameshort's focused inspection near critical moments versus uniform sampling's scattered temporal coverage, explaining the baseline's failure to detect the pink pen.
\end{itemize}

This case study yields two critical insights about \fancynameshort's temporal reasoning:
\begin{itemize}
\item \textbf{Sequential Event Disambiguation}: The system successfully differentiates between near-identical visual events through:
\begin{itemize}
\item First writing instance: Brown pen detection(false positive);
\item Second writing instance: Pink pen detection(true positive).
\end{itemize}

\item \textbf{Explanation of answer generation disparity}: \fancynameshort produces the correct answer (``\textit{Pink}'') versus uniform sampling's erroneous baseline (``\textit{Brown}'') due to temporal reasoning failures.
\end{itemize}

The spatial-temporal alignment between relevance peaks and ground truth positions confirms \fancynameshort's unique capacity to synchronize semantic logic with visual evidence flow. This case particularly highlights the method's superiority in scenarios requiring precise discrimination of recurrent events with subtle visual variations.

\section{Iteration Analysis}

\begin{figure}[htp]
\centering
\includegraphics[width=0.48\textwidth]{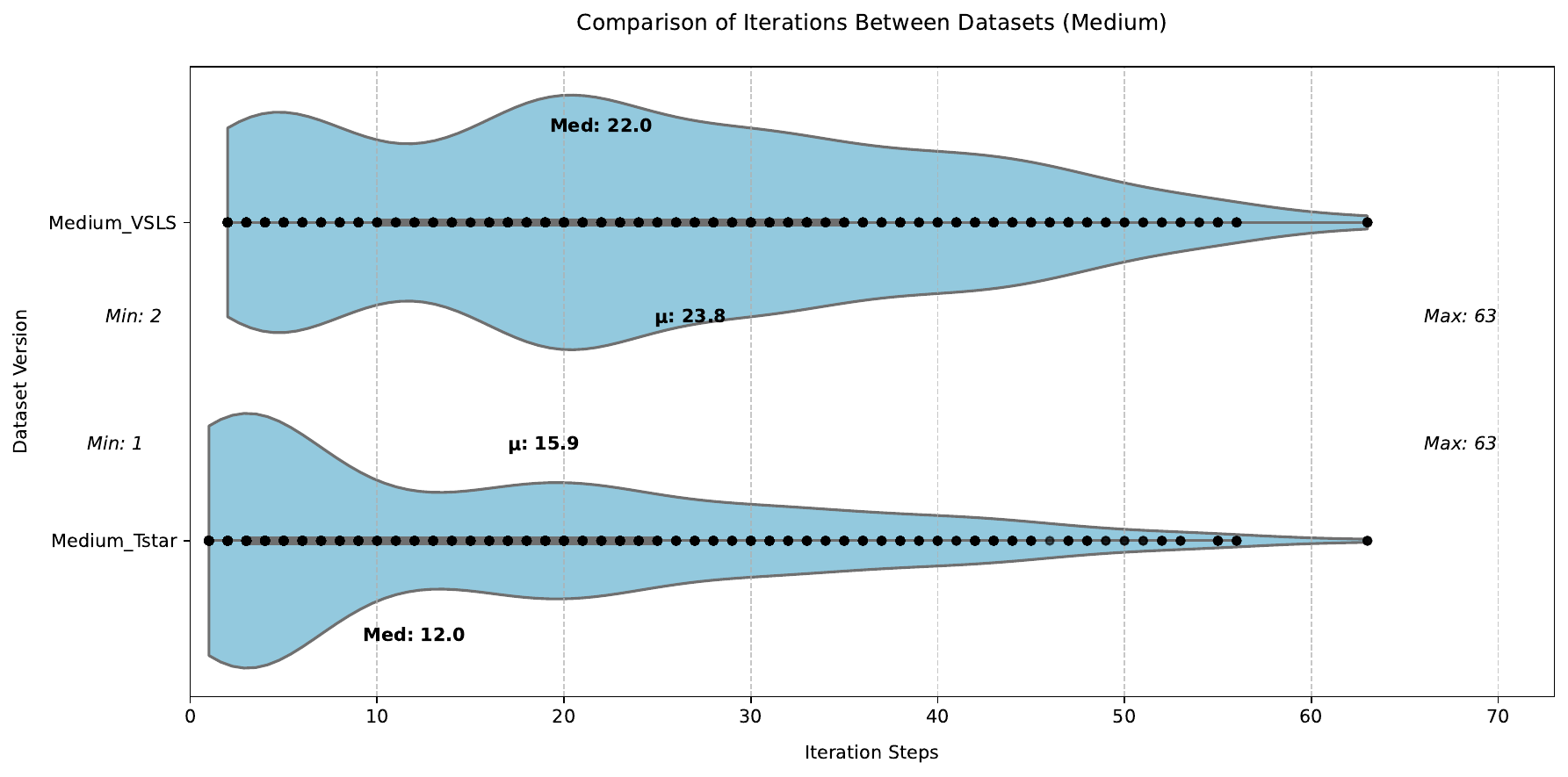}
%\vspace{-0.2 cm}
\caption{The comparative visualization of iteration counts on the medium-length video subset of the \videomme dataset demonstrates that our method consistently requires a higher number of iterations compared to the \tstar approach.
}
\label{violin}
%\vspace{-1 em}
\end{figure}

As shown in Fig~\ref{violin}, incorporating relations into the search algorithm will increase the average number of iterations for the video of medium length in the \videomme dataset from 15.9 to 23.8. The overall distribution of video iteration will not be significantly changed.

\clearpage

\section{Prompt}
\label{app:prompt}
\vspace{-0.5em}
\subsection{Prompt Template for Query Grounding}
\vspace{-0.5em}
Here is the prompt we used for query grounding.

\begin{minipage}{\textwidth}
\begin{tcolorbox}[
        standard jigsaw,
        title=Prompt Template for Query Grounding,
        opacityback=0,
        label=prompt_query_grounding,
        width=\textwidth,
    ]
Analyze the following video frames and the question:

Question: \texttt{<Question>}

Options: \texttt{<Options>} 

\textbf{Step 1}: Key Object Identification

\quad\textbullet{} Extract 3-5 core objects detectable by computer vision

\quad\textbullet{} Use YOLO-compatible noun phrases (e.g., “person”, “mic”)

\quad\textbullet{} Format: Key Objects: \texttt{obj1}, \texttt{obj2}, \texttt{obj3}

\textbf{Step 2}: Contextual Cues

\quad\textbullet{} List 2-4 scene elements that help locate key objects based on options provided

\quad\textbullet{} Use detectable items (avoid abstract concepts)

\quad\textbullet{} Format: Cue Objects: \texttt{cue1}, \texttt{cue2}, \texttt{cue3}

\textbf{Step 3}: Relationship Triplets

\quad\textbullet{} Relationship types:

\quad\quad\textbullet{} Spatial: Objects must appear in the same frame

\quad\quad\textbullet{} Attribute: Color/size/material descriptions (e.g., “red clothes”, “large”)

\quad\quad\textbullet{} Time: Appear in different frames within a few seconds

\quad\quad\textbullet{} Causal: There is a temporal order between the objects

\quad\textbullet{} Format of Relations: \texttt{(object, relation\_type, object)}, \texttt{relation\_type} should be exactly one of spatial/attribute/time/causal

\textbf{Output Rules}

\quad1. One line each for Key Objects/Cue Objects/Rel starting with exact prefixes

\quad2. Separate items with comma except for triplets where items are separated by semicolon
    
\quad3. Never use markdown or natural language explanations
    
\quad4.  If you cannot identify any key objects or cue objects from the video provided, please just identify the possible 
key or cue objects from the question and options provided 

\textbf{Below is an example of the procedure:}

\quad Question: For “When does the person in red clothes appear with the dog?”

\quad Response:
    
\quad\quad Key Objects: \texttt{person, dog, red clothes}
        
\quad\quad Cue Objects: \texttt{grassy\_area, leash, fence}
        
\quad\quad Rel: \texttt{(person; attribute; red clothes)}, \texttt{(person; spatial; dog)}

\textbf{Format your response EXACTLY like this in three lines:}

\quad\quad Key Objects: \texttt{object1, object2, object}
        
\quad\quad Cue Objects: \texttt{object1, object2, object}
        
\quad\quad Rel: \texttt{(object1; relation\_type1; object2)}, \texttt{(object3; relation\_type2; object4)}
\end{tcolorbox}
\end{minipage}

%\clearpage
\vspace{-0.5em}
\subsection{Prompt Template for Question Answering}
\vspace{-0.5em}
Here is the prompt we used for question answering.

\begin{minipage}{\textwidth}
\begin{tcolorbox}[
        standard jigsaw,
        title=Prompt Template for Question Answering,
        opacityback=0,
        label=prompt_qa,
        width=\columnwidth,
    ]
Select the best answer to the following multiple-choice question based on the video.

\texttt{<image>}

\texttt{<image>}

$\cdots$

Question: \texttt{<Question>}

Options: \texttt{<Options>}

Answer with the option’s letter from the given choices directly.

Your response format should be strictly an upper case letter A,B,C,D or E.
\end{tcolorbox}
\end{minipage}

\section{Limitations}
\label{app:limitations}
Despite the promising results of our \fancynameshort framework, we acknowledge several limitations: First, although our approach reduces the required frame sampling to just 1.4\%, the computational complexity remains a consideration for extremely long videos, with a search overhead of approximately 7.8 seconds. This may present challenges for real-time or low-latency applications. Besides, the performance of \fancynameshort is bounded by the capabilities of the underlying object detector (\textsc{YOLO-World}). Detection accuracy may degrade under challenging visual conditions such as poor lighting, occlusion, or unusual camera angles, potentially affecting temporal coverage.

\section{Broader Impacts}
\label{app:broader_impacts}
Our Visual Semantic-Logical Search (\fancynameshort) framework primarily offers positive societal impacts as a foundational algorithm for efficient keyframe selection in long videos.

\subsection{Positive Impacts}
\begin{itemize}
    \item \textbf{Educational Applications:} \fancynameshort enables students and educators to quickly locate relevant segments in instructional videos, improving learning efficiency for visual content.
    \item \textbf{Research Enhancement:} Scientists across disciplines can benefit from more efficient analysis of video archives, particularly those studying behavioral patterns or analyzing historical footage.
    \item \textbf{Computational Efficiency:} By sampling only 1.4\% of frames on average, our approach reduces computational requirements and energy consumption, contributing to more sustainable AI applications.
    \item \textbf{Accessibility:} Our framework can be integrated into assistive technologies for individuals with cognitive processing challenges, helping them identify and focus on critical moments in video content.
\end{itemize}

\subsection{Potential Considerations}
As a foundational algorithm, \fancynameshort has limited direct negative impacts. However, like any computer vision technology, applications built upon it should be mindful of general considerations:
\begin{itemize}
    \item \textbf{Underlying Model Biases:} The performance of \fancynameshort depends partly on object detection systems (e.g., YOLO-World), so it inherits any limitations or biases present in these components. Our modular design allows for substitution with improved detection systems as they become available.
\end{itemize}